\documentclass[10pt,twocolumn,letterpaper]{article}

\usepackage{iccv}
\usepackage{times}
\usepackage{epsfig}
\usepackage{graphicx}
\usepackage{amsmath}
\usepackage{amssymb}
\usepackage[toc,page]{appendix}
\usepackage{bibunits}
 \defaultbibliographystyle{ieee}
\usepackage{comment}
\usepackage{graphicx}
\usepackage{tabularx}
\usepackage{tabulary}
\newcommand*\samethanks[1][\value{footnote}]{\footnotemark[#1]}

\iccvfinalcopy 

\newcommand{\D}{D}
\newcommand{\ZZ}{K}
\newcommand{\B}{B}
\newcommand{\R}{R}
\newcommand{\xxpred}{\hat{\mathbf{x}}}
\newcommand{\xxgt}{\mathbf{x}^*}
\newcommand{\mdots}{..}

\DeclareMathOperator{\E}{\mathbb{E}}

\newcolumntype{K}[1]{>{\centering\arraybackslash}p{#1}}
\def\iccvPaperID{1967} 

\graphicspath{{figures/}}

\ificcvfinal\fi

\begin{document}
\begin{bibunit}[ieee]
\title{Adversarial Inverse Graphics Networks: Learning 2D-to-3D Lifting and Image-to-Image Translation from Unpaired Supervision}

\author{Hsiao-Yu Fish Tung \thanks{equal contribution} 
\qquad
Adam W. Harley \samethanks[1] 
\qquad
William Seto \samethanks[1] 
\qquad
Katerina Fragkiadaki
\\
Carnegie Mellon University \\
{\tt\small \{htung,aharley,wseto,katef\}@cs.cmu.edu}
}

\maketitle


\begin{abstract}

Researchers have developed excellent feed-forward models that learn to map images to desired outputs, such as to the images' latent factors, or to other images, using supervised learning. Learning such mappings from unlabelled data, or improving upon supervised models by exploiting unlabelled data, remains elusive. We argue that there are two important parts to learning without annotations: 
(i) matching the predictions to the input observations, and (ii) matching the predictions to known priors.
We propose Adversarial Inverse Graphics networks (AIGNs): weakly supervised neural network models that combine feedback from rendering their predictions, with distribution matching between their predictions and a collection of ground-truth factors. We apply AIGNs to 3D human pose estimation and 3D structure and egomotion estimation, and outperform  models supervised by only paired annotations. We further apply AIGNs to facial image transformation using super-resolution and inpainting renderers, while deliberately adding biases in the ground-truth datasets. Our model seamlessly incorporates such biases, rendering input faces towards young, old, feminine, masculine or Tom Cruise-like equivalents (depending on the chosen bias), or adding lip and nose augmentations while inpainting concealed lips and noses.


\end{abstract}

\begin{figure}[t!]
    \centering
    \includegraphics[width=\linewidth]{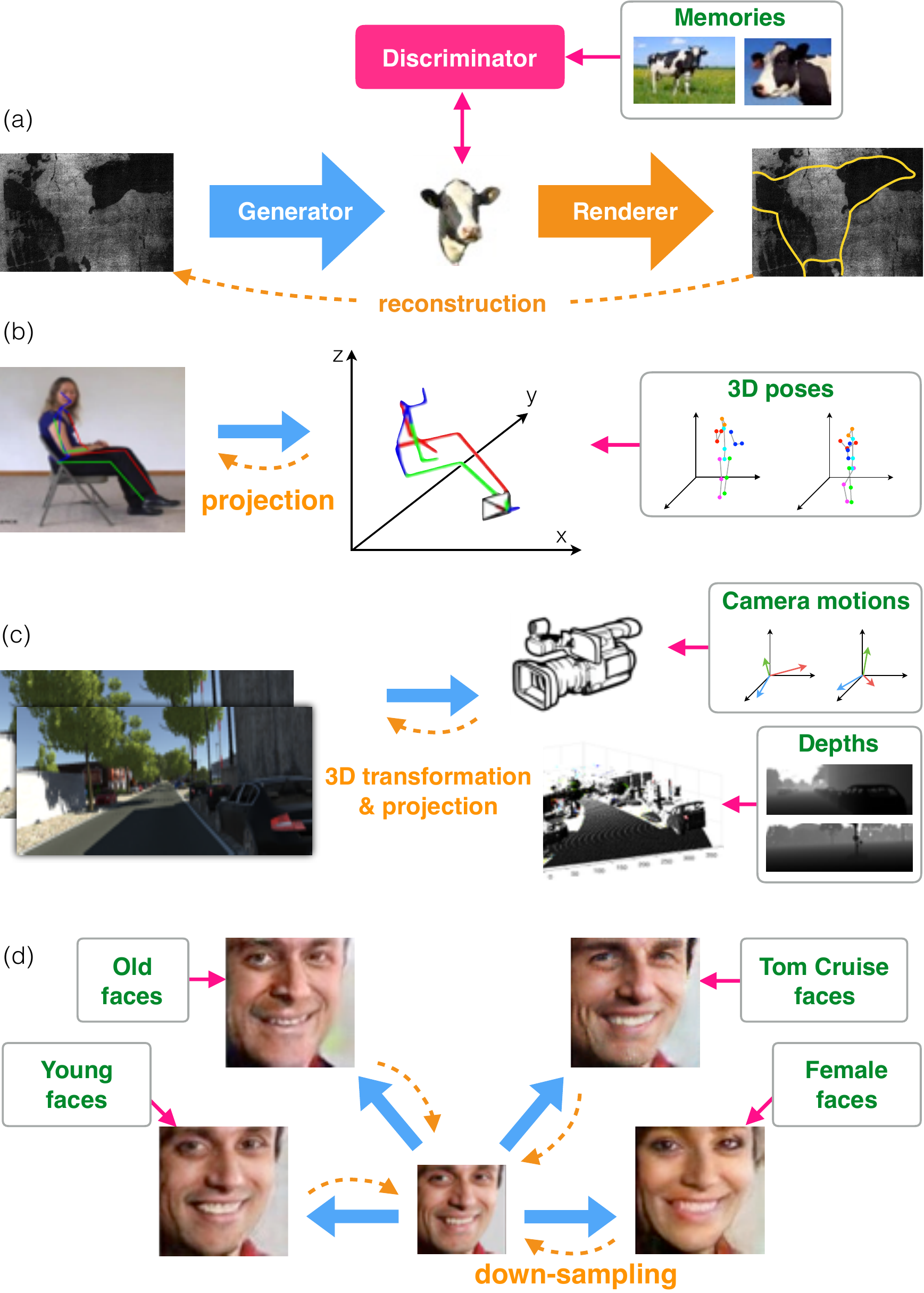}
     \centering
    \caption{\textbf{Adversarial Inverse Graphics Networks} combine feedback from differentiable rendering of the predictions, with priors, imposed through distribution matching between predictions and ``memories". The general architecture is shown (a) with Dallenbach's classic cow illusion \cite{cow}, and subsequently with our experimental tasks: (b) 3D human pose estimation, (c) extraction of 3D depth and camera motion from a pair of frames, and (d) creative and controllable image generation. Each task uses a unique domain-specific renderer, depicted in orange dashed arrows. 
    }
\label{fig:architecture}
\end{figure}

\section{Introduction}
Humans imagine far more than they see. 
In Figure~\ref{fig:architecture}~(b), we imagine the hidden arms and legs of the sitting woman. In Figure~\ref{fig:architecture}~(c), we imagine forward motion of the camera, as opposed to the road drifting backwards underneath the camera. 
We arrive at these interpretations -- \ie, predictions of latent factors -- by referring to \textit{priors} on how the world works. 

\begin{figure}[t!]
    \centering
    \includegraphics[width=\linewidth]{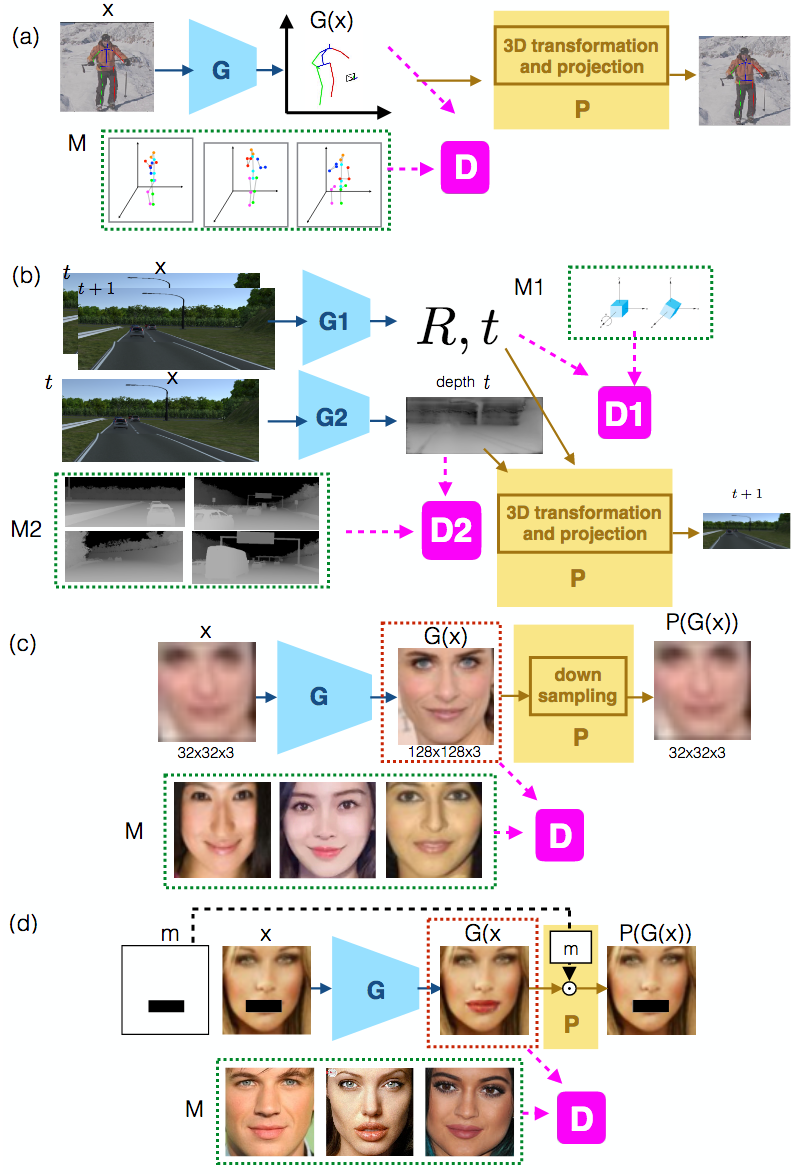}
     \centering
    \caption{\textbf{Adversarial inverse graphics architectures} for (a) 3D human pose estimation, (b) Structure from motion, (c) super-resolution, and (d) inpainting. }
\label{fig:framework}
\end{figure}

This work proposes Adversarial Inverse Graphics Networks (AIGNs),  a model that learns to map images to latent factors using feedback from the rendering of its predictions, as well as distribution matching between the predictions and a stored collection of ground truth latent factors. 
The ground-truth collection does not need to be directly related to the current input -- it can be a disordered set of labels. We call this \textit{unpaired supervision}. The renderers employed are differentiable, parameter-free, and task specific, \eg, camera projection, camera motion, downsampling, and masking. 
Figure~\ref{fig:framework} depicts architectures for AIGNs for the tasks of structure from motion, 3D human pose estimation, super-resolution, and in-painting. 

When we (deliberately) \textit{bias the ground truth} collection of an AIGN to reflect a distorted reality, surprising renders arise. 
For super-resolution, AIGNs can make people look older, younger, more feminine, more masculine or more like Tom Cruise, by simply curating the discriminators' ground truth to include images of old, young, female, male, or Tom Cruise pictures, respectively. 
For inpainting, AIGNs can make people appear with bigger lips or bigger noses, 
again by curating the discriminators' ground truth to include faces with big lips or big noses. These observations inspire a compelling analogy: the AIGN's ground truth collection is its \textit{memory}, and the renders are its \textit{imagination}. When the memories are biased, the imaginations reflect a distorted reality.

Our model is related to, and builds upon, many recent works in the literature. Inverse-graphics networks \cite{DBLP:journals/corr/KulkarniWKT15} use \textit{parametric} deconvolutional networks and strong supervision (\ie, annotation of images with their ground truth imaginations) for decomposing an   image into interpretable factors, \eg, albedo and shading.  
Our model instead employs \textit{parameter-free} renderers, and makes use of unpaired weak supervision. 
Similar to our approach, 3D interpreter networks \cite{Wu2016} use a reconstruction loss in the form of a 2D re-projection error of predicted 3D keypoints, along with paired  supervised training for 3D human pose estimation.  
Our model complements the reconstruction loss with adversarial losses on the predicted 3D human poses, and shows performance superior to the 3D interpreter. 
Conditional Generative Adversarial Networks (GANs) have used a combination of adversarial and L2 losses, \eg, for  inpainting \cite{pathakCVPR16context}, 3D voxel generation \cite{3dgan},  super-resolution \cite{DBLP:journals/corr/DongLHT15}, and image-to-image translation \cite{DBLP:journals/corr/IsolaZZE16}. In these models, the adversarial loss is used to avert the regression-to-mean problem in standard L2 regression (\ie blurring). 
Such models are feed-forward; they do not have a renderer and reconstruction feedback. 
As a result, they can only exploit supervision in the form of annotated pairs \eg, of an RGB image and its corresponding 3D voxel grid \cite{3dgan}. 
Our model extends such supervised conditional GAN formulations through  self-consistency via inverse rendering, and matching  between \textit{rendered predictions and the visual input}, rather than between predictions and the ground truth. This feedback loop allows weakly supervised distribution matching to work (using \textit{unpaired} annotations for the discriminators), removing the need for direct label matching. 

As unsupervised models, AIGNs do not discriminate between training and test phases. 
Extensive experiments in a variety of tasks show that their performance consistently improves over their supervised alternatives, by adapting in a self-supervised manner to the statistics of the test set.


\section{Adversarial Inverse Graphics Networks} 

AIGN architectures for various tasks 
are shown in Figure \ref{fig:framework}. 
Given an image or image pair $x$, 
 generator networks $G_i, \ i \in 1 \mdots \ZZ$ map $x$ to  a set of predictions  $ \mathcal{G}(x) = \{G_i(x), i \in 1 \mdots \ZZ \}$. 
 Then, a task-specific differentiable renderer $P(\mathcal{G}(x))$ renders predictions  back to the original input space.  Discriminator networks $\D_i, i \in 1 \mdots \ZZ$ are trained to discriminate between predictions $G_i(x)$ and  true \textit{memories} of appropriate form $M_i$. 
 Discriminators have to assign large probability to  true memories  and low probability to generators' predictions. 
 Given a set of images (or image pairs depending on the task) $X = \{x_1, x_2, \mdots, x_n\},$ 
 generators are trained to minimize L2 distance between rendered imaginations and input $x$ and simultaneously maximize discriminators' confusion. 
Our loss  then reads:


\begin{align}
\label{eq:model}
   \min_{\mathcal{G}} \max_{\mathcal{D}} & \E_{x \in X} 
 \underbrace{\| P(\mathcal{G}(x)) - x \| _2}_{\text{reconstruction loss}} + \nonumber \\
&\beta \sum \limits_{i=1}^\ZZ  \underbrace{\log \D_i(M_i) + \log(1-\D_i(G_i(x)))}_{\text{adversarial loss}}, 
\end{align}
 where $\D$ denotes the set of discriminator networks and  $\mathcal{G}$ the set of generator networks and $\beta$ the relative weight of reconstruction and adversarial losses. Since paired ground-truth is not used anywhere, both reconstruction and adversarial losses can be used both at train and test time. Our model can also benefit from strong supervision through paired annotations -pairs of visual input with desired predictions- at training time for training the generator networks. We will use the term \textit{adversarial priors} to denote adversarial losses over latent predicted factors of our model.

In the next sections, we present applications of AIGNs to the tasks of (i) 3D human pose estimation, (ii) extraction of 3D depth and egomotion from a pair of frames, which we call Structure from Motion (SfM),  (iii) super-resolution, and  (iv) inpainting. 
For  application of AIGNs to intrinsic image decomposition, please see our earlier unpublished manuscript \cite{inversionearlier}. 


\subsection{3D Human Pose Estimation}
Recent works use deep neural networks and large supervised training sets \cite{h36m_pami} and learn to regress to 3D poses directly, given an RGB image \cite{DBLP:journals/corr/PavlakosZDD16}. Many have explored 2D body pose as an intermediate representation \cite{DBLP:journals/corr/ChenR16a,Wu2016,DBLP:journals/corr/TomeRA17}, or as an  auxiliary task in a multi-task setting \cite{DBLP:journals/corr/TomeRA17} where the abundance of labelled 2D pose training examples helps feature learning and complements limited 3D supervision. Yasin \etal \cite{Yasin_Iqbal_CVPR2016} explored unpaired supervision between 2D and 3D keypoint annotations by pretraining a low-rank Gaussian model from 3D annotations as a prior for plausible 3D poses. Instead of fitting 3D poses into a predefined probabilistic model and having a separate stage for pretraining, our model learns such priors directly from data, and co-trains the networks responsible for the priors and the predictions.

The AIGN architecture for 3D human pose estimation is depicted in Figure \ref{fig:framework} (a). 
Given an image crop centered around a person detection, the task is to  
predict a 3D skeleton for the depicted person.   
We decompose the 3D human shape into a view-aligned 3D linear basis model 
and a rotation matrix: 
$x_{3D}=\R\cdot \sum_{j=1}^{|\mathcal{B}|}\alpha_j \B_j$
Our shape basis $\mathcal{B}$ is obtained using PCA on orientation-aligned 3D poses in our training set, where orientation is measured by the direction of the normal to the hip vector. 
We keep 60 components out of a total of 96 (three coordinates of 32 keypoints).
The dimensionality reduction is small, and indeed, we just use bases weights for ease of prediction, relying on our adversarial priors (rather than PCA) to regularize the 3D shape prediction. 

Our generator network first predicts a set of 2D bodyjoint heatmaps 
and then maps those to 
basis weights $\mathbf{w} \in \mathbb{R}^{60}$, focal length $f$, and  Euler rotation angles $\alpha,\beta,\gamma$ so that the 3D rotation of the shape reads $\R=\R^x(\alpha)\R^y(\beta)\R_t^z(\gamma)$, where $\R^x(\theta)$ denotes rotation around the x-axis by angle $\theta$.   
Our renderer then projects the 3D keypoints $x_{3D}$ to 2D keypoints $x^{proj}_{2D}$, all in homogeneous coordinates: 
\begin{equation}
x^{proj}_{2D}=P \cdot x_{3D} + \begin{bmatrix} c_x \\c_y \\ 0 \end{bmatrix},
\end{equation}
where $P=\begin{bmatrix} f & 0 & 0 & 0 \\ 0 & f & 0 & 0 \\ 0 & 0 & 1 & 0 \end{bmatrix}$ is the camera projection matrix. 
The reconstruction loss  is the $L2$  norm between the reprojected  coordinates $x^{proj}_{2D}$ and 2D coordinates obtained by the argmax of predicted 2D heatmaps.  
A discriminator network discriminates between our generated keypoints $x_{3D}$ and a database of 3D human skeletons, which does not contain the ground-truth (paired) 3D skeleton for the input image, but rather previous poses (\ie, ``memories'').

\subsection{Structure from Motion}

Simultaneous Localization And Mapping (SLAM)  methods have shown impressive results on estimating camera pose and 3D point clouds from monocular, stereo, or RGB-D video sequences \cite{schoeps14ismar,kerl13iros} using alternate optimization over camera poses and the 3D scene pointcloud. 
There has been a recent interest in integrating learning to aid  geometric approaches handle moving objects and low-texture image regions. The work of Handa et al. \cite{handa2016gvnn} contributed a  deep learning library with many geometric operations including a differentiable camera projection layer, similar to those used in SfM networks \cite{sfmnet,tinghuisfm}, 3D image interpreters \cite{interpreter}, and  depth from stereo  CNN~\cite{garg2016unsupervised}. Our SfM AIGN architecture, depicted in Figure \ref{fig:framework} (b), build upon those works. Given a pair of consecutive frames, the task is to predict the relative camera motion and a  depth map for the first frame.

We have two generator networks: an egomotion network and a depth network. The egomotion network takes the following inputs: a concatenation of two consecutive frames $I_1,I_2$, their 2D optical flow estimated using our implementation of FlowNet \cite{flownet}, the angles of the optical flow, and a static angle field. The angle field denotes each pixel's angle from the camera's principal point (in radians), while the flow angles denote the angular component of the optical flow. The egomotion network produces the camera's 3D relative rotation and translation  $\{ R , T \} \in SE(3)$ as output. We represent 3D camera relative rotation using an Euler angle representation:
$\R=\R^x(\alpha)\R^y(\beta)\R_t^z(\gamma)$ where $\R^x(\theta)$ denotes rotation around the $x$-axis by the angle $\theta$. 
The egomotion network is a standard convolutional architecture (similar to VGG16 \cite{simonyan2014very}), but following the last convolution layer, there is a single fully-connected layer producing a 6D vector, representing the Euler angles of the rotation matrix, and the translation vector. 

The depth network  takes as input the first frame $I_1$ and estimates 3D scene log depth at every pixel, $\{ d \} \in \mathbb{R}^{w\times h}$. The architecture of this network is the same as that of the egomotion network, except deconvolution layers (with skip connections) replace the fully-connected layer, producing a depth estimate at every pixel. 
 Given generated depth $d$ for the first frame $I_1$ and known camera intrinsics -- focal length $f$ and principal point $(c_x,c_y)$ -- we  obtain the corresponding 3D point cloud for $I_1$:  
$\mathbf{X}_1^i = \begin{bmatrix} X_1^i \\ Y_1^i \\ Z_1^i \end{bmatrix} =
\frac{d^i}{f}\begin{bmatrix}
x_1^i - c_x \\
y_1^i - c_y\\
f \\
\end{bmatrix}$,
where $(x_1^i, y_1^i)$ are the column and row  coordinates and $d^i$ the predicted depth of the $i^{th}$ pixel in frame $I_1$. 
Our renderer in this task transforms the point cloud $\mathbf{X}_1$ according to the estimated camera motion: 
$\mathbf{X}_2 = \R \cdot \mathbf{X}_1 + T  $ 
(similar to SE3-Nets \cite{SE3-Nets} and the recent SfM-Nets \cite{sfmnet,tinghuisfm}), and projects the 3D points back to image pixels $x_2,y_2$ using  camera projection. 

The optical flow vector of pixel $i$  is then $(U^i,V^i)=(x^i_{2} - x^i_1, y^i_{2} - y^i_1)$.  Our reconstruction loss corresponds to  the well known brightness constancy principle, that pixel appearance does not change under small pixel motion:
\begin{equation}
   \mathcal{L}^{photo} = \frac{1}{w~h}\sum_{x,y} \|I_1(x, y) - I_2(x + U(x, y), y + V(x, y)) \|_1
\end{equation}

We have two discriminator networks: $\D_1$ manages the realism of predicted camera motions by enforcing physics constraints in a statistical  manner: when the camera is fixed to the top of a car, the roll and pitch angles are close to zero due to the nature of the car motion. This network has three fully-connected layers (of size 128, 128, and 64), and discriminates between real and fake relative transformation matrices. 
The second discriminator, $\D_2$, 
manages the realism of 3D depth predictions, using a collection of ground-truth depth maps from unrelated sequences. This discriminator is \textit{fully convolutional}. That is, it has one probabilistic output at each image sub-region, as we are interested in the realism of local depth texture. 
We found the depth discriminator to effectively handle the scale ambiguity of monocular 3D reconstruction: in a monocular setup without priors, the re-projection error cannot tell if an object is far and big, or near and small. The depth discriminator effectively imposes a prior regarding world scale, as well as depth realism on fine-grained details.  

\subsection{Super-resolution}
Deep neural network architectures have recently shown excellent performance on the task of super-resolution \cite{DBLP:journals/corr/DongLHT15,DBLP:journals/corr/BrunaSL15} and non-blind inpainting \cite{pathakCVPR16context,DBLP:journals/corr/YehCLHD16} (inpainting where the mask is always at the same part of the image). Adversarial losses have been used to resolve the regression to the mean problem of standard $L_2$ loss \cite{DBLP:journals/corr/LedigTHCATTWS16,pathakCVPR16context}. 
Our model unifies recent works of \cite{sonderby2014apparent} and \cite{DBLP:journals/corr/YehCLHD16} that combine adversarial and reconstruction losses for super-resolution and inpainting respectively, without paired supervision, same as our model. 
However, weak (unpaired) supervision might be unnecessary for  super-resolution or inpainting  as an unlimited amount of ground-truth pairs can be easily collected by downsampling or masking images. Our AIGN focuses instead on \textit{biased} super-resolution and  inpainting, as a tool for creative and controllable image generation. 

The AIGN architecture for super-resolution is depicted in Figure \ref{fig:framework} (c). 
Our generator network takes as input a low-resolution image $x \in \mathbb{R}^{w \times h}$ and produces a high resolution image $G(x)\in \mathbb{R}^{4W \times 4H}$ after a number of residual neural blocks \cite{he2016deep}.  Our renderer $P(G(x))$  is a downsampler that reduces the size of the output image by four times. We implement it using average pooling with the appropriate stride. Our reconstruction loss is the $L2$ distance \textit{between the input image and the rendered imagination $P(G(x))$}. 
Our discriminator  takes as input high resolution predicted images, as well as memories of high resolution images, unrelated to the current input.  Thus far, our model is similar to the concurrent work S{\o}nderby \etal \cite{sonderby2014apparent}.  

Unsupervised super-resolution networks \cite{sonderby2014apparent} may not be necessary, since  large supervised training datasets of low- and high-resolution image pairs can be collected by Gaussian blurring and downsampling.   
We instead focus on what we call \textit{biased super-resolution} for face images. We  bias our discriminator ground-truth memories to contain high resolution images of a particular image category, \eg, females, males, young, old faces, or faces of a particular individual. AIGNs  then mix the low-frequencies of the input image (preserved via our reconstruction loss) with high frequencies from the memories (imposed by the adversarial losses), the relative weight $\beta$ between reconstruction and adversarial loss in Eq. \ref{eq:model} controls such mixing.  The result is realistically looking  faces whose age, gender or identity has been  transformed, as shown in Figure  \ref{fig:superresolution}. 
Such facial transformations are completely unsupervised; the model has never seen a pair of the same person old and young (or male and female). 

\begin{table*}[t]
\centering
\begin{tabular}{K{0.12\textwidth}K{0.04\textwidth}K{0.04\textwidth}K{0.03\textwidth}K{0.03\textwidth}K{0.03\textwidth}K{0.04\textwidth}K{0.03\textwidth}K{0.05\textwidth}K{0.04\textwidth}K{0.05\textwidth}K{0.04\textwidth}K{0.03\textwidth}K{0.03\textwidth}K{0.05\textwidth}}
 \hline
  & Direct & Discuss & Eat & Greet & Phone & Photo & Pose & Purchase & Sit & SitDown & Smoke & Wait& Walk &  Average\\
 \hline
  Forward2Dto3D & 75.2 & 118.4 & 165.7 & 95.9 & 149.1 & 154.1 & 77.7 & 176.9 & 186.5 & 193.7 & 142.7 & 99.8 &  74.7 &  128.9\\
  3Dinterpr \cite{Wu2016} & 56.3 & 77.5 & 96.2 & 71.6 & 96.3 & 106.7 & 59.1 & 109.2 &111.9 & {\bf 111.9} & 124.2 & 93.3 & 58.0  & 88.6 \\
    Monocap \cite{DBLP:journals/corr/ZhouZPLDD17}&  78.0 & 78.9 & 88.1 & 93.9 & 102.1 & 115.7 & 71.0 & {\bf 90.6} &  121.0 & 118.2 & 102.5 & 82.6 &  75.62 &  92.3  \\
  AIGN (ours)  & {\bf 53.7} & {\bf 71.5} & {\bf 82.3} & {\bf 58.6} & {\bf 86.9} & {\bf 98.4} & {\bf 57.6} & {\bf 104.2} & {\bf 100.0} & 112.5 & {\bf 83.3} & {\bf 68.9} & {\bf 57.0} &  {\bf 79.0} \\
 \hline
\end{tabular}
\vspace{1mm}
\caption{\textbf{3D reconstruction error} in H3.6M using ground-truth 2D keypoints as input.}\label{tab:tab1}
\end{table*}

\begin{table*}[t]
\centering
\begin{tabular}{K{0.12\textwidth}K{0.04\textwidth}K{0.04\textwidth}K{0.03\textwidth}K{0.03\textwidth}K{0.03\textwidth}K{0.04\textwidth}K{0.03\textwidth}K{0.05\textwidth}K{0.04\textwidth}K{0.05\textwidth}K{0.04\textwidth}K{0.03\textwidth}K{0.03\textwidth}K{0.05\textwidth}}
 \hline
  & Direct & Discuss & Eat & Greet & Phone & Photo & Pose & Purchase & Sit & SitDown & Smoke & Wait& Walk &  Average\\
  \hline
  Forward2Dto3D & 80.2 & 92.4 & 102.8 & 92.5 & 115.5 & 79.9 & 119.5 & 136.7 & 136.7& 144.4 & 109.3 & 94.2 & 80.2 & 104.6 \\
  3Dinterpr \cite{Wu2016} &  78.6 & {\bf 90.8} & 92.5 & 89.4 & 108.9 & 112.4 &  77.1 & {\bf 106.7} & 127.4 & 139.0 & 103.4 & 91.4 &  79.1 &  98.4 \\
  AIGN (ours) & {\bf 77.6} &  91.4 &  {\bf 89.9} & {\bf 88} & {\bf 107.3 }& {\bf 110.1 }&  {\bf 75.9} &  107.5 & {\bf 124.2} & {\bf 137.8} & {\bf 102.2} & {\bf 90.3} &{\bf  78.6} &  {\bf 97.2} \\ 
 \hline
\end{tabular}
\vspace{1mm}
\caption{\textbf{3D reconstruction error} in H3.6M using detected 2D keypoints as input.}\label{tab:tab2}
\end{table*}

\subsection{Inpainting}
The AIGN architecture for inpainting is depicted in Figure \ref{fig:framework} (d). 
The input is a ``masked" image $x$, that is, an image whose content is covered by a black contiguent mask $m$. Our generator produces a complete (inpainted) image $G(x)$. 
The rendering function  $P$ in this case is 
a masking operation: $P(G(x),m) = m \odot G(x),$ where $\odot$ denotes pointwise multiplication. Our discriminator receives inpainted imaginations $G(x)$ and memories of complete face images $M$, unrelated to our current input images. Our model is trained then to predict complete, inpainted  images that when masked will match the input image $x$. Thus far, our model is similar to  \cite{DBLP:journals/corr/YehCLHD16}.

Unsupervised inpainting networks \cite{DBLP:journals/corr/YehCLHD16}  may not be necessary, since large  supervised training datasets of paired masked and complete images can be  collected via image masking.     
We instead focus on \textit{biased inpainting} of face images.  
We bias the discriminator's ground-truth memories to contain complete images with a  particular desired characteristic, in the location of the mask $m$. For example, if the mask covers the mouth of a person, we  can bias discriminators' ground-truth memories  to only contain people with big lips. In this case, our generator will produce inpainted  images $G(x)$, that have this localized characteristic in a smooth photorealistic blend, in order to confuse the discriminator $\D$.

For further details on the proposed architectures please see the supplementary material.

\section{Experiments}


\subsection{3D human pose estimation from a static image}

We use the Human3.6M (H3.6M) dataset of Ionescu \etal \cite{h36m_pami}, the largest available dataset with annotated 3D human poses. This dataset contains videos of actors performing activities and provides annotations of body joint locations in 2D and 3D at every frame, recorded from a Vicon system. We split the videos by the human subjects, with five subjects (S1, S5, S6, S7, S8) for training
and two subjects (S9, S11) for testing, following the split of previous works  \cite{DBLP:journals/corr/ZhouZPLDD17}.
For both sets, we use one third of the original frame rate.

We consider a variety of supervised setups and baselines for our 3D human pose predictor, which we detail below. 
We first train our network using synthetic data augmentation (Sup 1), following the protocol of Wu \etal \cite{Wu2016}: A 3D example skeleton is first perturbed,  a 3D rotation $\R$ and focal length $f$ are sampled, and the resulting rotated shape is projected to 2D. 
We further train our network using real paired 2D-to-3D training data from H3.6M  (Sup 2). Our generator network trained with Sup1+Sup2 we will call it \textit{Forward2DTo3D} net, 
 as it resembles a standard supervised model for 3D human pose estimation.  
We further 
finetune using a reconstruction loss (2D reprojection error) in the test data (Sup 3). Our generator network trained with Sup1+Sup2+Sup3 we will call it \textit{3D interpreter} due to its clear similarity with Wu \etal \cite{Wu2016}. Since the original source code is not available, we re-implement it and use it as one of our baselines. 
Our AIGN model, along with Sup1+Sup2+Sup3, uses unsupervised adversarial losses in the test data using randomly selected 3D  training poses (Sup4). 
We compare AIGN to 3D interpreter and Forward2Dto3D baselines in two setups for 3D lifting: (a) using ground-truth 2D body joints provided by H3.6M as input, and (b) using 2D body joints provided by the state-of-the-art Convolutional Pose Machine detector \cite{wei2016cpm}. We used an SVM regressor to map keypoint definitions  of Wei \etal \cite{wei2016cpm} to the one of H3.6M dataset.  When using ground-truth 2D body joints as input we also compare against the publicly available 3D pose code for  MonoCap \cite{DBLP:journals/corr/ZhouZPLDD17}, an optimization method that uses  a sparsity prior across an over-complete dictionary of 3D poses, and minimizes the reprojection error via Expectation-Maximization. We consider one image as input for all the models for a fair comparison (MonoCap was originally proposed assuming a video sequence as input). 

\textbf{Evaluation metrics.}
Given a set of estimated 3D joint locations $\xxpred_1 \cdots \xxpred_K$ and  corresponding ground-truth 3D joint locations $\xxgt_1 \cdots \xxgt_K$, 
the reconstruction error is defined as the 3D per-joint
error after the torsos are aligned to face the front with transformation $\mathcal{T}$
:$\frac{1}{K}\sum_{i=1}^K \| \mathcal{T}(\xxpred_i)- \mathcal{T}(\xxgt_i) \|.$
We show the 3D reconstruction error (in millimeters) of our model and baselines in Tables~\ref{tab:tab1} and \ref{tab:tab2}, organized according to activity, following the presentation format of Zhou \etal \cite{DBLP:journals/corr/ZhouZPLDD17}, though \textit{only one model was used across all activities} for our method and baselines (for MonoCap this means using the same dictionary for optimization in all images).   

The AIGN outperforms the baselines, especially for ground-truth keypoints. This suggests it would be valuable to finetune 2D keypoint detector features as well, instead of keeping them fixed. 
Adversarial priors allow the model not to diverge when finetuned on new (unlabelled) data, as they ensure anthropomorphism and plausibility of the detected poses.  For additional 3D pose results, please see the supplementary material.

\begin{figure} [h!]
    \centering
    \includegraphics[width=1.0\linewidth]{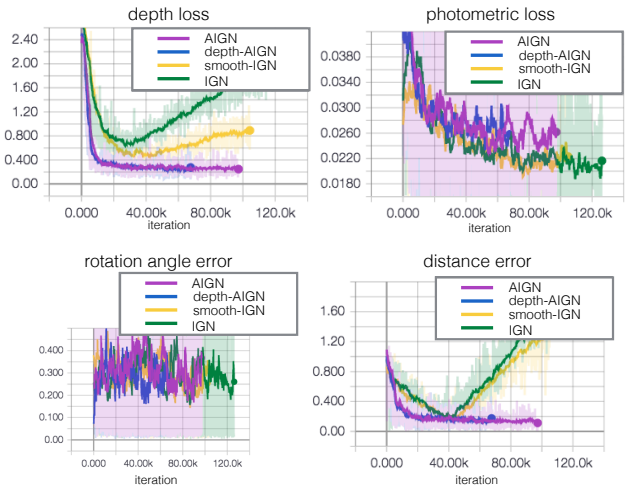}
     \centering
    \caption{Depth, camera motion (relative rotation and translation) and  photometric reprojection error curves during training of (a) the AIGN model with adversarial losses on both depth and camera motion (\textit{AIGN}), (b) an AIGN with adversarial losses only on depth (\textit{depth-AIGN}), (c) a model with reconstruction loss and depth smoothness loss (\textit{smooth-IGN}) and (d) a model with a reconstruction loss only (\textit{IGN}). Adversarial priors handle scale ambiguity of reconstruction loss, and thus models do not diverge.}
    \label{fig:training-sfm-difficult}
\end{figure}

\begin{figure*} [h!]
    \centering
    \includegraphics[width=1.0\linewidth]{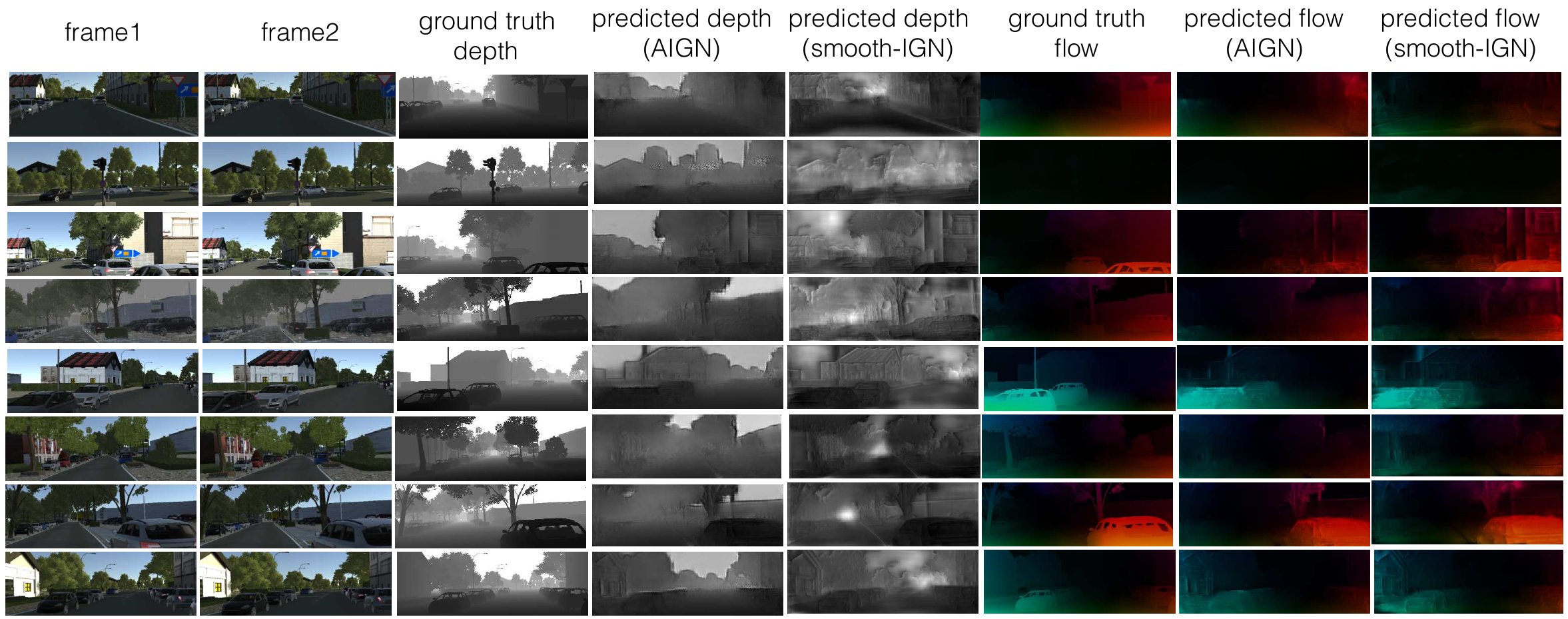}
     \centering
    \caption{\textbf{Structure from motion results with and without adversarial priors.} The results of the baseline (columns $5$th and $8$th) are obtained from a model with depth smoothness prior (\textit{smooth-IGN}), trained with early stopping at 40K iterations (before divergence).}
    \label{fig:sfm-visual}
\end{figure*}

\subsection{Structure from Motion}
We use Virtual KITTI (VKITTI) \cite{Gaidon:Virtual:CVPR2016}, a synthetic dataset that depicts videos taken from a camera mounted on a car driving in an urban center. This dataset contains scenes and camera viewpoints similar to the real KITTI dataset \cite{Geiger2012CVPR}. We use the VKITTI dataset rather than KITTI, because the real-world dataset has large errors in the ground truth camera motion for short sub-sequences, whereas the synthetic ground truth is precise. 
We use the first VKITTI sequence (in all weather conditions) as the validation set, and the remaining four sequences as the training set.
We consider the tasks of (i) single-frame depth estimation, and (ii) egomotion estimation from pairs of frames. 

\textbf{Evaluation metrics.} We evaluate the error between our camera motion predictions using four metrics, as defined in prior works \cite{sturm12iros, DBLP:conf/icra/JaeglePD16}. These are: (a) distance error: the camera end point error distance in meters; (b) rotation angle error: the camera rotational error in radians, (c) angular translation error: the error in the angular direction of the camera translation; and (d) magnitude translation error: the error
in the magnitude of the camera translation. 

We evaluate the error between our depth prediction and the true depth with an L1 distance in log depth space. The use of log depth is inspired by Eigen \etal \cite{eigen2014depth}, but we do not use the scale-invariant error from that work, because it assumes the presence of an oracle indicating the true mean depth value at test time. 

We consider two supervision setups: (a) unsupervised learning of SfM, where the AIGN is trained from scratch using  (self-supervised) reconstruction and adversarial losses, and (b) supervised pretraining,  where depth and camera motion in the training set are used to pretrain our generators, before applying unsupervised learning in the test set.

\textbf{Self-supervised learning of SfM.}
We compare  the following models: (a) our full model, \textit{AIGN}, with adversarial losses on depth and camera motion, (b) our model with adversarial losses only on depth, but not on camera motion (\textit{depth-AIGN}), (c) our model without adversarial losses but instead with a smoothness loss on depth (\textit{smooth-IGN}), and (d) our model without any adversarial priors (\textit{IGN}). 


We show depth, camera error and photometric errors against number of training iterations for all models in Figure \ref{fig:training-sfm-difficult} in the test set.  Models without depth adversarial priors diverge after a number of iterations (the depth map takes very large values).  This is due to the well known \textit{scale ambiguity} of monocular reconstruction: objects can be further away or simply be smaller in size, and the 2D re-projection will be the same. Adversarial priors enforce constraints regarding the general scale of the world scene, as well as depth-like appearance and camera motion-like plausibility on the predictions, and prevent depth divergence. While the adversarial model has higher  photometric error that the one without priors (naturally, since it is more constrained), the intermediate variables, namely depth and camera motion, are much more accurate when adversarial priors are present. The model that uses depth smoothness (\textit{depth-IGN}) falls in-between the AIGN model and the model without any priors, and diverges as well. In Figure \ref{fig:sfm-visual}, we show the estimated depth and geometric flow predicted by models with and without adversarial priors.

\textbf{Supervised pretraining.}
In this setup, we pretrain our camera pose estimation sub-network and depth sub-network using the VKITTI training set using  supervised training against ground-truth depthmaps and camera motions supplied with the dataset (\textit{pretrain} baseline).  We then  finetune our model in the test set using self-supervision, \ie, reprojection error and adversarial constraints  (\textit{Pretrain+SelfSup}). 
We  evaluate camera pose estimation performance of our models in the test set, and compare against the geometric model of Jaegle \etal \cite{DBLP:conf/icra/JaeglePD16}, a monocular camera motion estimation method that takes optical flow as input, and solves for the camera motion with an optimization-based method that attenuates the influence of outlier flows. 
We show  our translation and rotation camera motion errors in Table \ref{tab:SfM}.  The pretrained model performs worse than the geometric baseline. When pretraining is combined with self-supervision, we obtain a much lower error than the geometric model.  Monocular geometric methods, such as \cite{DBLP:conf/icra/JaeglePD16}, do not have a good way to resolve the scale ambiguity in reconstruction, and thus have large translation errors. The AIGN for SfM, while being monocular, learns depth and does not suffer from such ambiguities. Further, we outperform \cite{DBLP:conf/icra/JaeglePD16} with respect to the angle of translation that a geometric method can in principle estimate (no ambiguity). 
These results suggest that our adversarial SfM model  improves by simply watching unlabeled videos, without diverging. 

\begin{table}[h]
\centering
\begin{tabular}{c c c c c}
 \hline
 & trl err & rot err & trl mag. & trl ang.  \\
 \hline
 Geometric \cite{DBLP:conf/icra/JaeglePD16} & 0.4588 & \textbf{0.0014} & 0.4579 & 0.3423 \\
 Pretrain  & 0.4876 & 0.0017 & 0.4865 & \textbf{0.3306} \\
 Pret.+SelfSup & \textbf{0.1294} & \textbf{0.0014} & \textbf{0.1247} & 0.3333 \\
 \hline
\end{tabular}
\vspace{1mm}
\caption{\textbf{Camera motion estimation in our Virtual KITTI test set}. The self-supervised model outperforms the geometric baseline of Jaegle \etal \cite{DBLP:conf/icra/JaeglePD16}. 
The translation error (column 1) is decomposed into magnitude and angular error in columns 3-4.  } \label{tab:SfM}
\end{table}

\subsection{Image-to-image translation}
We use the CelebA dataset \cite{liu2015faceattributes} 
which contains 202,599 face images, with 10,177 unique identities, and is annotated with 40 binary attributes.  We preprocess the data by  cropping each face image to the largest bounding box that includes the whole face using the OpenFace library \cite{amos2016openface}. 

\begin{figure}[h!]
\begin{tabular}{c}
\small{(a) Female-to-male transformation.} \\
\vspace{2mm}
\includegraphics[width=0.98\linewidth]{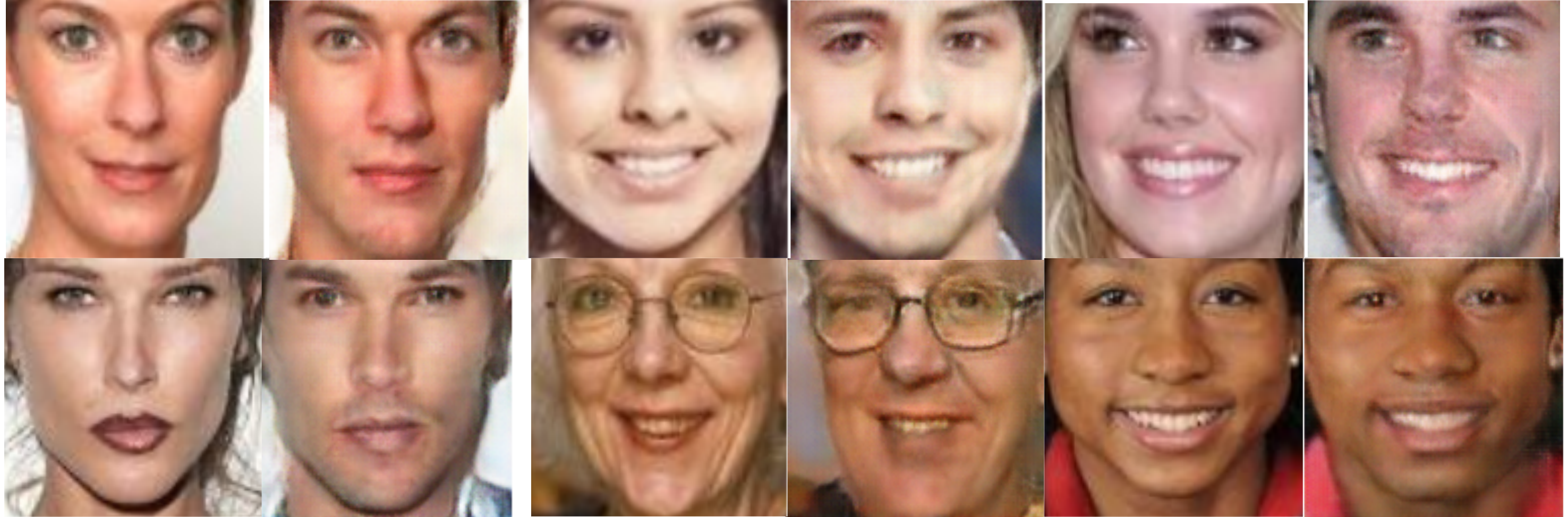} \\
\small{(b) Male-to-female transformation.} \\
\vspace{2mm}
\includegraphics[width=0.98\linewidth]{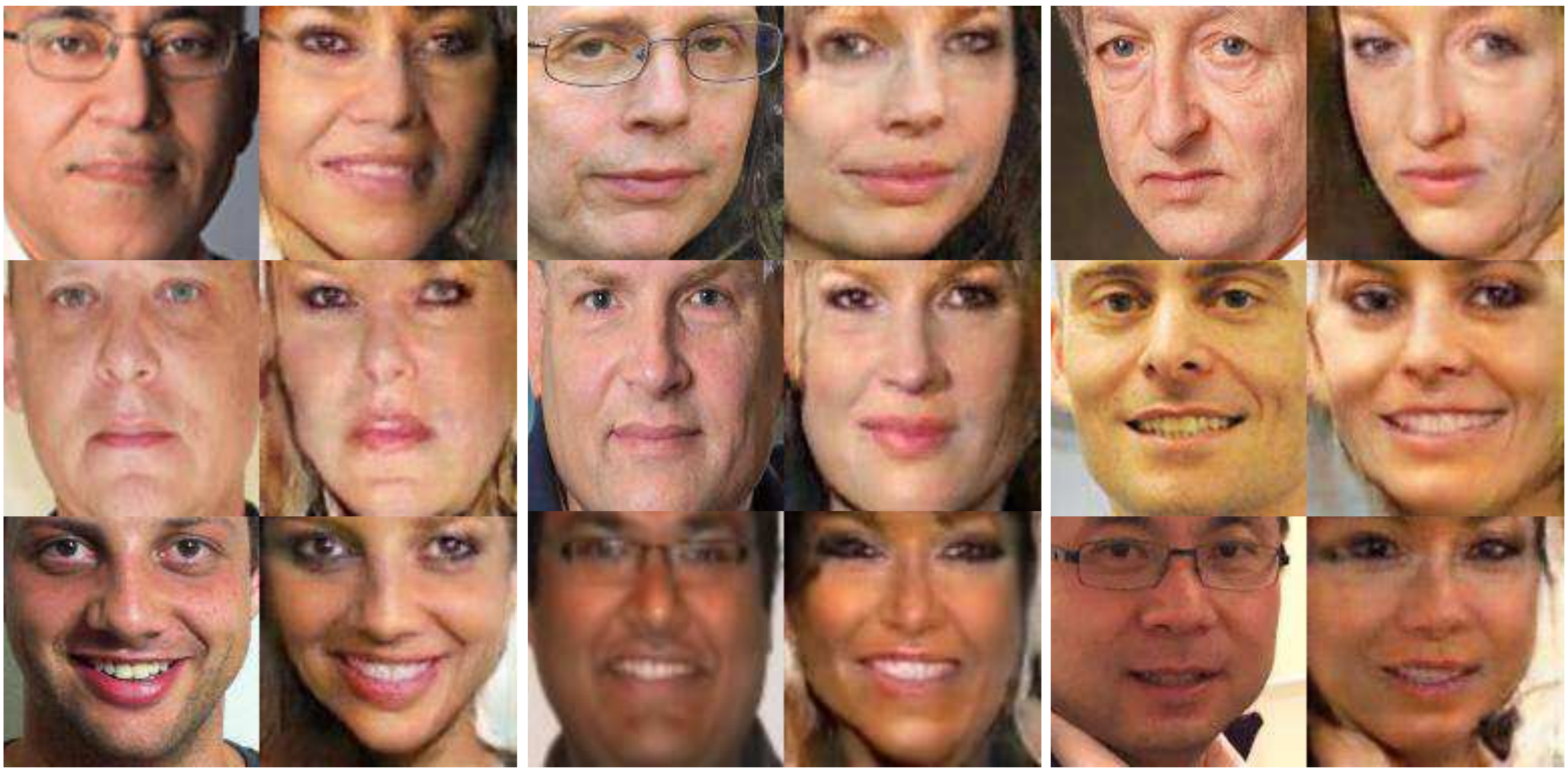} \\
\small{(c) Anybody-to-Tom Cruise transformation.} \\
\vspace{2mm}
\includegraphics[width=\linewidth]{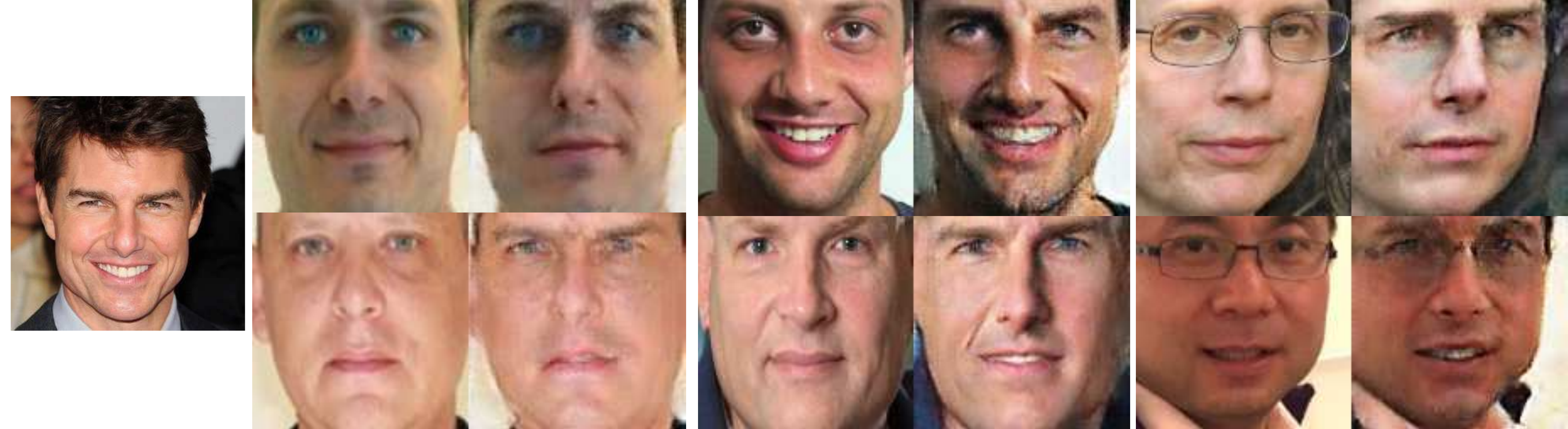} \\
\small{(d) Comparison with Attribute2Image \cite{DBLP:journals/corr/YanYSL15} for male-to-female}\\
\small{(left) and old-to-young transformation (right).} \\
\vspace{2mm}
\includegraphics[width=\linewidth]{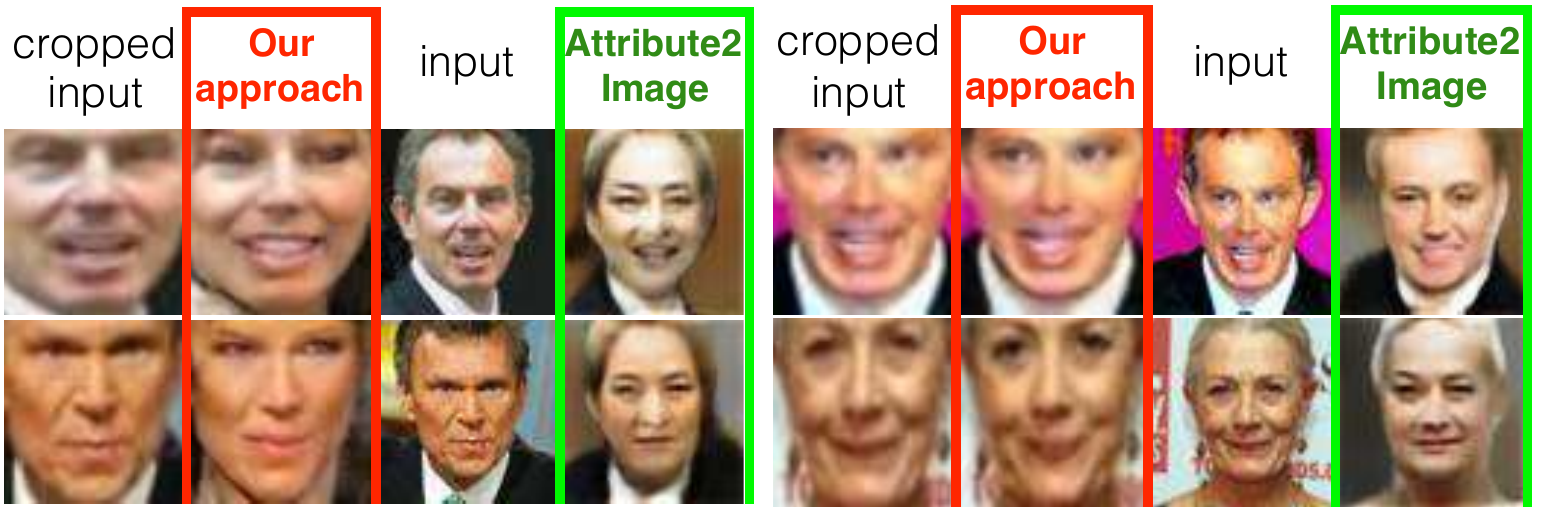} \\
\end{tabular}
\centering
\caption{\textbf{Biased adversarial super-resolution}.}\label{fig:superresolution}
\end{figure}

\textbf{Biased super-resolution.} We train female-to-male and male-to-female gender  transformation by applying  adversarial super-resolution to new face input images, while discriminator memories contain only male or only female faces,  respectively. We train old-to-young and young-to-old age   transformations by applying  adversarial super-resolution to new face images while discriminator memories contain only young or only old faces,  respectively -- as indicated by the attributes in the CelebA dataset. 
We train identity mixing transformations by applying  adversarial super-resolution to new face images while discriminator memories contain only a particular person identity, for demonstration we choose Tom Cruise. We show results in Figure \ref{fig:superresolution} (a-c). 
We further compare our model against the recent work of  Attribute2Image\cite{DBLP:journals/corr/YanYSL15} in Figure \ref{fig:superresolution}(d) using code available by the authors. The AIGN better preserves the fidelity of the transformation and is more visually detailed. Though we demonstrate age, gender transformation and identity mixing, AIGN could be used for any creative image generation task, with appropriate curation of the discriminator's ground-truth memories. 

\begin{figure}
    \centering
    \includegraphics[width=0.97\linewidth]{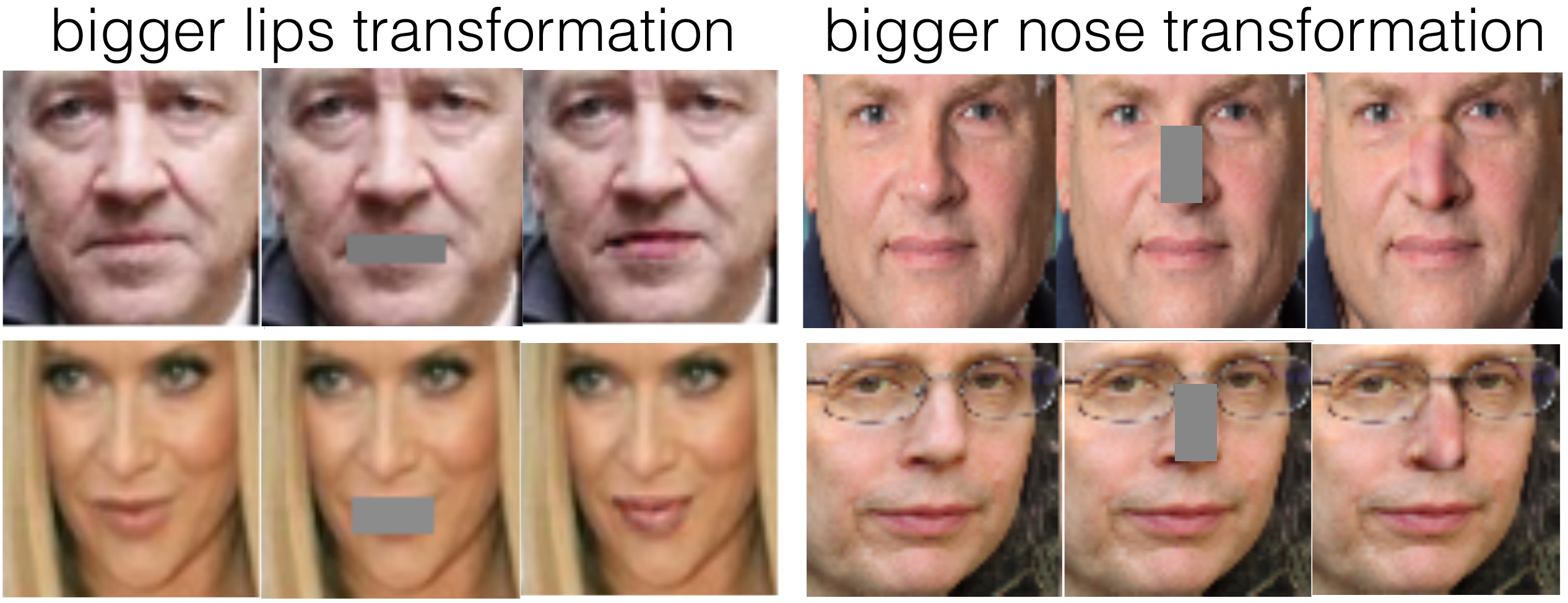}
     \centering
    \caption{\textbf{Biased adversarial inpainting} for bigger lips transformation (left) and bigger nose transformation (right). From left to right: the original image, the masked image input to our model, and the generated output superimposed over the masked image.}
    \label{fig:inpainting}
\end{figure}

\textbf{Biased Inpainting.} We train ``bigger lips" and ``bigger nose" transformations by applying adversarial inpainting to new input face images where the mouth or nose region has been masked, respectively, and discriminator's memories contain only face images with big lips or with  big noses, respectively. Note that ``big lips" and ``big nose" are attributes annotated in the CelebA dataset. We show results in Figure \ref{fig:inpainting}. 
From top to bottom, we show the original image, the  masked image input to adversarial inpainting, the output of our generator, and in the last row, the output of our generator superimposed over the original image.

\begin{figure} [h!]
    \centering
    \includegraphics[width=\linewidth]{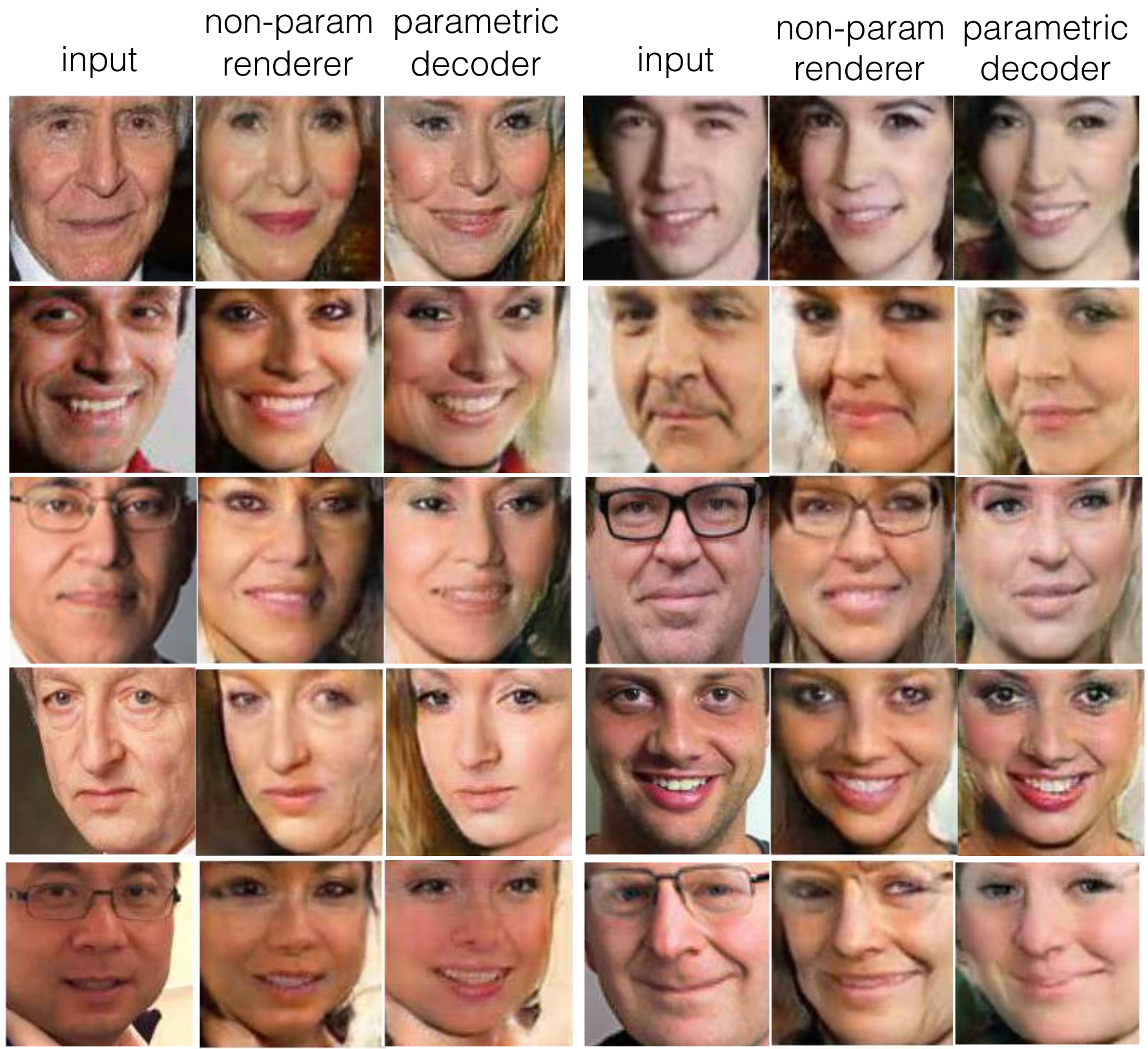}
     \centering
    \caption{Comparing male-to-female transformation using  renderers and parametric decoder.}
    \label{fig:para-nonpara}
\end{figure}

\textbf{Renderers versus parametric decoders for image-to-image translation.} 
We compare the results of domain-specific, non-parametric renderers, and parametric decoders in  image-to-image translation tasks.  
For our model with parametric decoder we use the full res image as input and instead of the downsampling module in the proposed super-resolution renderer
we instead use convolutional/deconvolutional layers so that the decoder  can freely adjust its weights through training. 
This is similar to the one way transformation proposed in the concurrent work of \cite{pmlr-v70-kim17a} \cite{CycleGAN2017}. 
We trained both models on gender transformation, and  results are shown 
in Figure \ref{fig:para-nonpara}. While both models produce photorealistic results, the model with the renderer produces females ``more paired" to their male counterparts, while parametric renderers may alter other properties of the face considerably, e.g., in the last row of Figure \ref{fig:para-nonpara}, the age of the produced females does not match their male origins. Parameter-free rendering is an important element of unsupervised learning with AIGNs; We have observed that  parametric decoders (instead of parameter-free renderers) can  cause the reconstruction loss to drop without learning  meaningful predictions but rather exploiting the capacity of the decoder. We provide a comprehensive experiment in the supplementary material.

\section{Conclusion}
We have presented Adversarial Inverse Graphics Networks, weakly supervised neural networks for 2D-to-3D lifting and image-to-image translation that combines feedback from  renderings of predictions with data-driven priors on  latent semantic factors, imposed using adversarial networks. 
We showed AIGNs outperform  previous supervised models that do not use adversarial priors, in the tasks of 3D human pose estimation and  3D structure and egomotion extraction. We further showed how biasing discriminators' priors  for inpainting and super-resolution results in creative image editing, and outperforms supervised variational autoencoder models of previous works, in terms of the fidelity of the transformation, and the amount of visual detail. Deep neural networks have shown that we do not need to engineer our features; AIGNs shows we do not need to engineer our priors either.
\putbib[refs]
\end{bibunit}

\clearpage
\appendix
\begin{bibunit}[ieee]
\begin{appendices}

\section{Parametric vs. non-parametric decoders}
Here we discuss the benefits of using non-parametric and domain-specific renderers, over learned decoders.
Both the proposed model and CycleGAN \cite{CycleGAN2017} can be viewed as autoencoders: the input is first transformed 
into a target domain, and then transformed back to its original space.
A parametric decoder could be more desirable, for the reason that we do not need to hand-engineer a mapping function from the target domain back to the inputs.
However, simply using reconstruction loss and adversarial loss does not
guarantee that the predictions look spatially similar to the inputs. In tasks such as image-to-image translation, spatial precision can be of critical importance.
With a parametric decoder, the transformed input can be viewed as a information bottleneck, and as long as the decoder can correctly ``guess'' the final output from the transformed input (\ie, the code), the code is valid and the solution is optimal.

To support this point, we conduct an experiment on image inpainting using the MNIST dataset. 
Similar to the parametric encoder-decoder described in the main text, the network has two main parts:
(1) an encoder that transforms the input (a partially obscured image of a digit) into prediction (a hallucinated digit), 
and (2) a decoder
that transforms the prediction back into the input. 
Instead of using convolutional layers, which have an architectural bias on 
preserving spatial relationships, we use fully-connected layers in both the 
encoder and the decoder. This is important, because such architectural conveniences are unavailable in less-structured tasks, such as 3D pose prediction and SfM. 
We train the model with a reconstruction loss on the decoder, and adversarial loss on the encoder.

\begin{figure}[t]
    \centering
    \includegraphics[width=\linewidth]{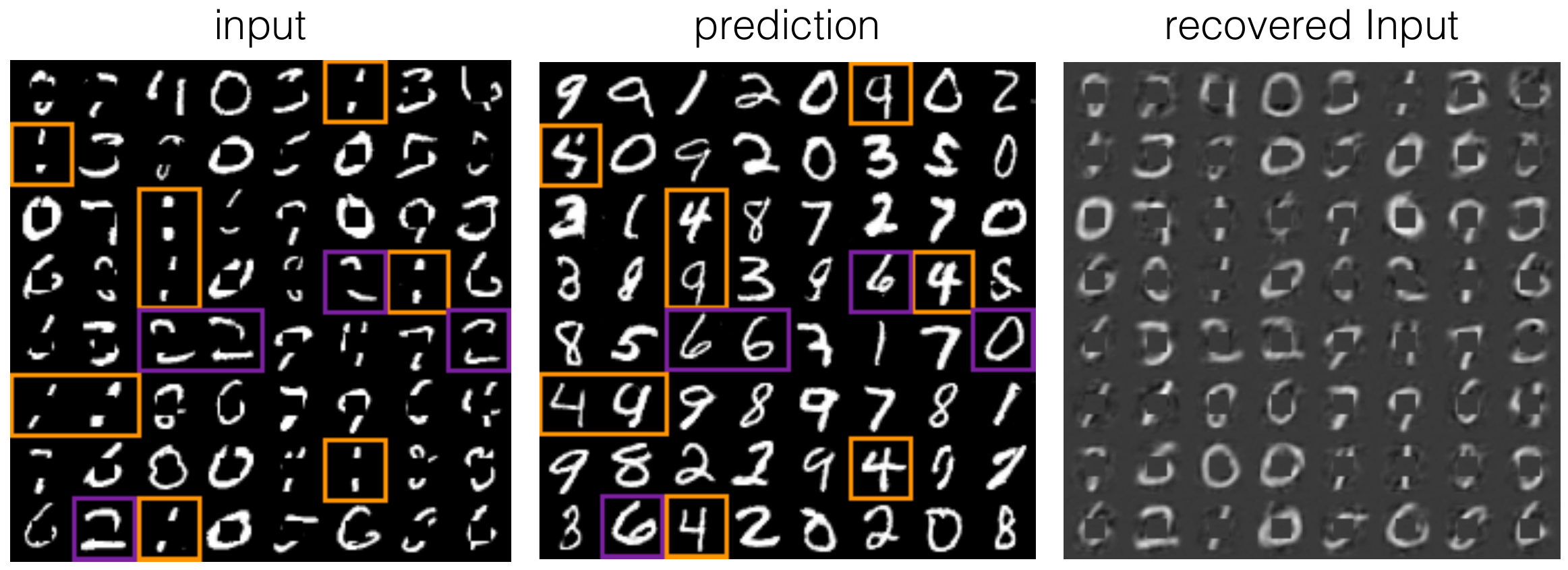}
     \centering
    \caption{Digit inpainting using an encoder-decoder architecture with fully-connected layers. Many predictions are incorrect, while the recovered inputs are accurate. Orange squares highlight instances of the digit ``1'' transformed into other digits; purple squares highlight instances of the digit ``2'' transformed into other digits.}
    \label{fig:mnist}
\end{figure}

The results are shown in Figure \ref{fig:mnist}.
While inpainting, the encoder (incorrectly) transforms many of the digits into other digits. For instance, several obscured ``1'' images are inpainted as ``4''. In the parametric decoding process, however, these errors are \textit{undone}, and the original input is recovered successfully. In other words, the decoder takes the burden of the reconstruction loss, allowing the encoder to learn an inaccurate latent space. Parameter-free rendering avoids this problem.

\section{Additional experiments and details}
In the sections to follow,  we provide implementation details, including architecture descriptions for the generator and discriminator in each task, and training details. Additionally, we provide more experimental results.

\subsection{3D human pose estimation from static images}
\begin{figure}[t]
    \begin{tabular}{l}
    \includegraphics[width=0.9\linewidth]{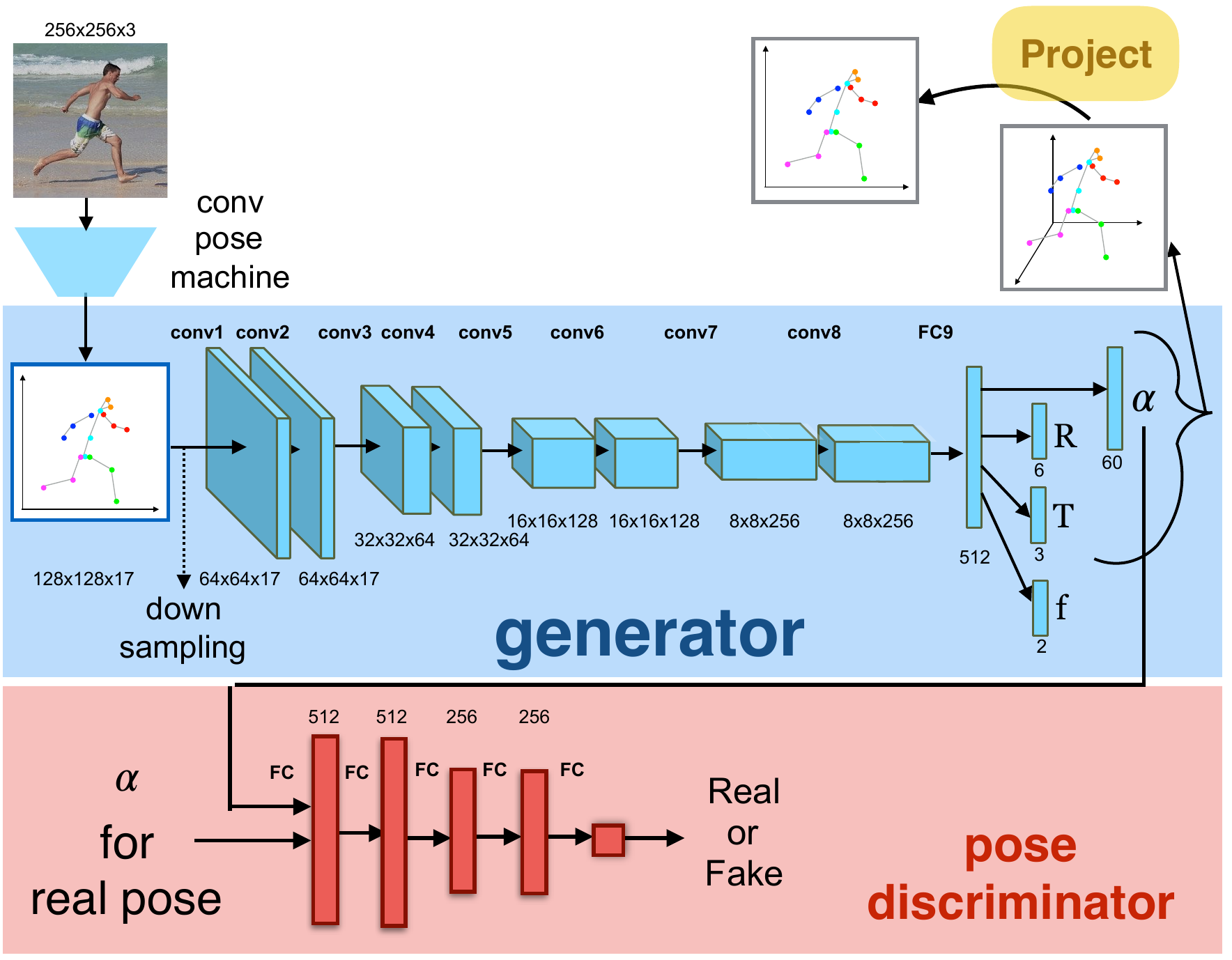} \\
    \end{tabular}
\centering
    \label{fig:architectures1}
    \caption{\textbf{Generators and discriminators' architectures for the task of 3D human pose estimation from a single image.}}
\end{figure}

\begin{figure}[t]
    \begin{tabular}{l}
    \includegraphics[width=0.9\linewidth]{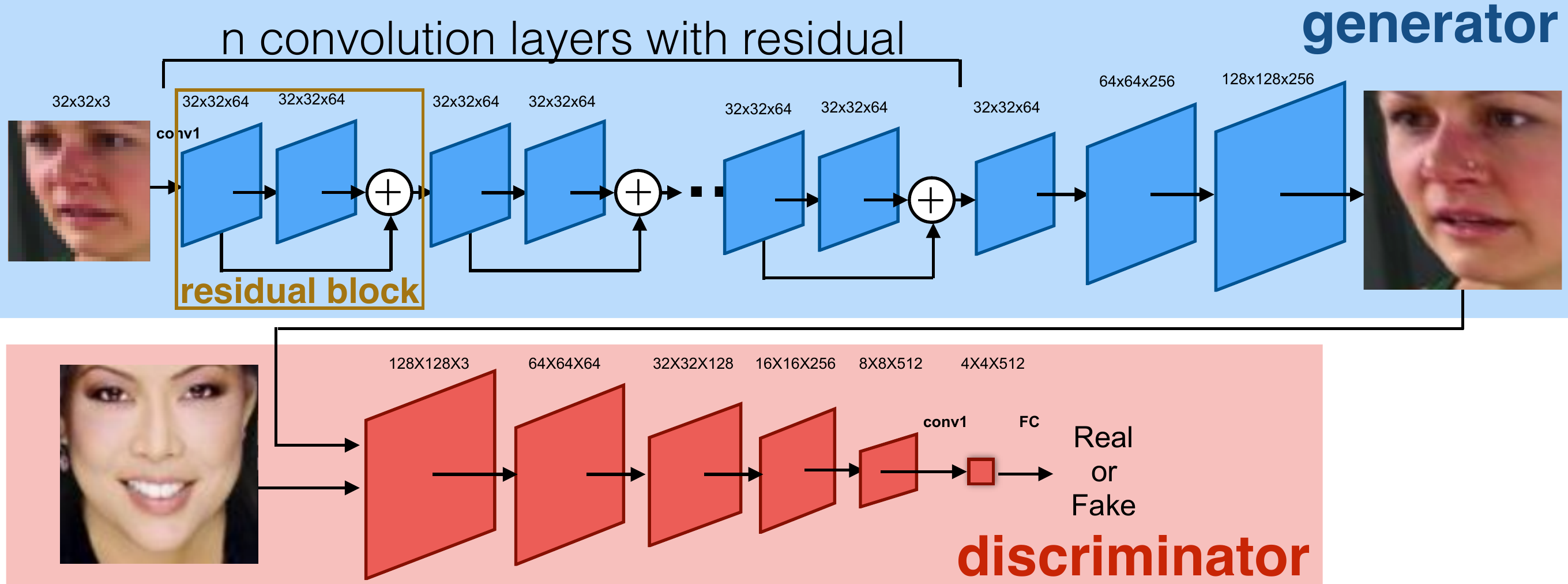} \\
    (a) Generator's and discriminator's architectures for  \\image super-resolution. \\
    \includegraphics[width=0.9\linewidth]{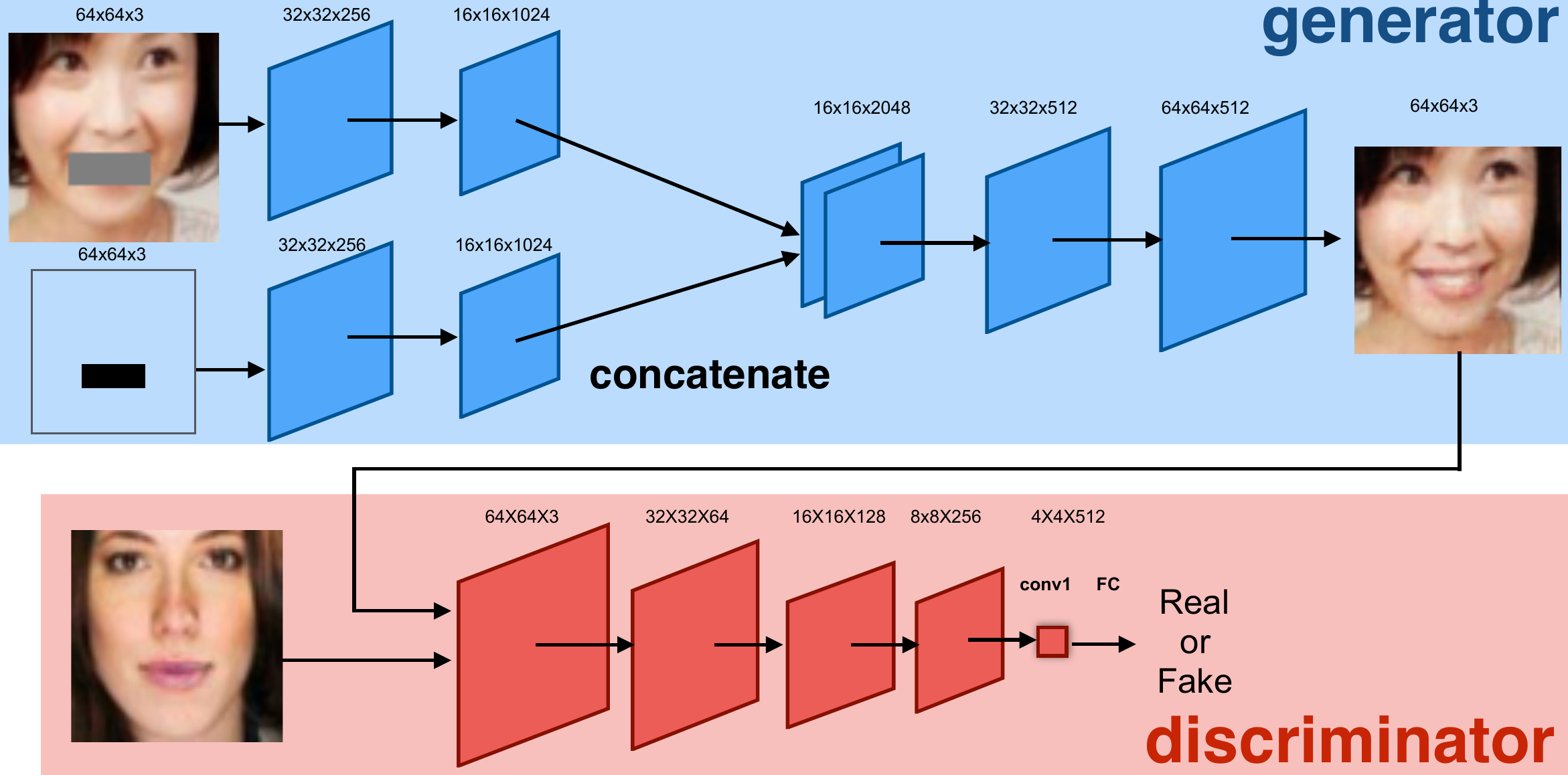} \\
    (b) Generator and discriminator architectures for image  \\inpainting. \\
    \end{tabular}
\centering
    \caption{\textbf{Architectures for AIGN.}}
    \label{fig:architectures}
\end{figure}

Figure~\ref{fig:architectures1} shows the architecture of our generator network for 3D human pose estimation from a single  RGB image. Our generator predicts 
weights over the shape bases $\alpha$, rotation $R,$ translation $T$ and focal length $f$, as described in our paper. 
The generator takes as input a set of 2D body joint heatmaps. 
We use convolutional pose machines \cite{wei2016cpm} 
to estimate 2D keypoints, and convert them into heatmaps  by creating a Gaussian distribution  
centered 
around each 2D keypoint. 
The network consists of 8 convolutional layers with leaky ReLU activations and batch normalization and two fully connected layers at the end that map to the desired outputs. The width, height and number of channels for each layer is specified in Figure~\ref{fig:architectures}. 
The discriminator for this task consists of five fully connected layers with featuremap depth 512, 512, 256, 256 and 1, with a leaky ReLU and batch normalization after each layer. The discriminator takes all values output from the generator (\ie, $\alpha$, $R,$ $T$,  $f$) as input. 

In all experiments, we set the variance for the Gaussian heatmap $\sigma$ to 0.25, 
and the dimensionality of our PCA shape basis to 60 (out of 96 total bases). The dimensionality reduction is small, and indeed, we only use basis weights for ease of prediction, relying on our adversarial priors (rather than PCA) to regularize the 3D shape prediction. We use gradient descent for both generator and discriminator training. Learning rate for reconstruction loss is set to 0.00001  and learning rate for the adversarial loss is set to 0.0001. All parameters are initialized with random sampling from zero mean normal distributions with variance of 0.02.

\begin{figure}[t]
    \centering
    \includegraphics[width=0.9\linewidth]{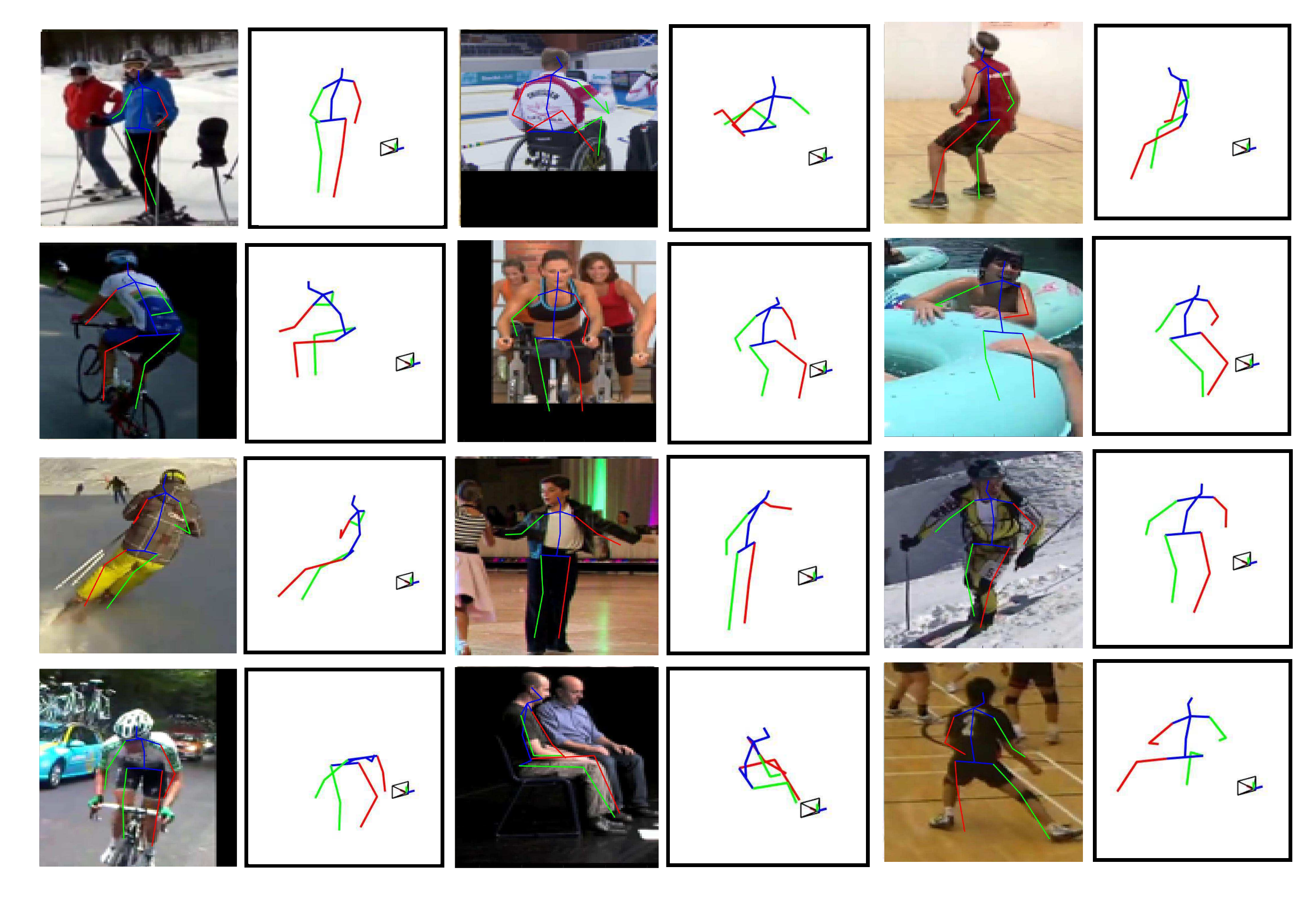}
     \centering
    \caption{\textbf{Predicted 3D human poses in MPII dataset using the supplied ground-truth 2D keypoints as input.} }
    \label{fig:mpii}
\end{figure}

In Figure \ref{fig:mpii}, we show predicted 3D human poses on  images from the MPII dataset \cite{andriluka14cvpr} using the ground-truth 2D keypoints available. Our model generalizes well \textit{on unseen  images without any further self-supervised finetuning}, though we would expect  additional self-supervised finetuning to further improve performance. 

\begin{figure}[t]
    \begin{tabular}{l}
    \includegraphics[width=0.9\linewidth]{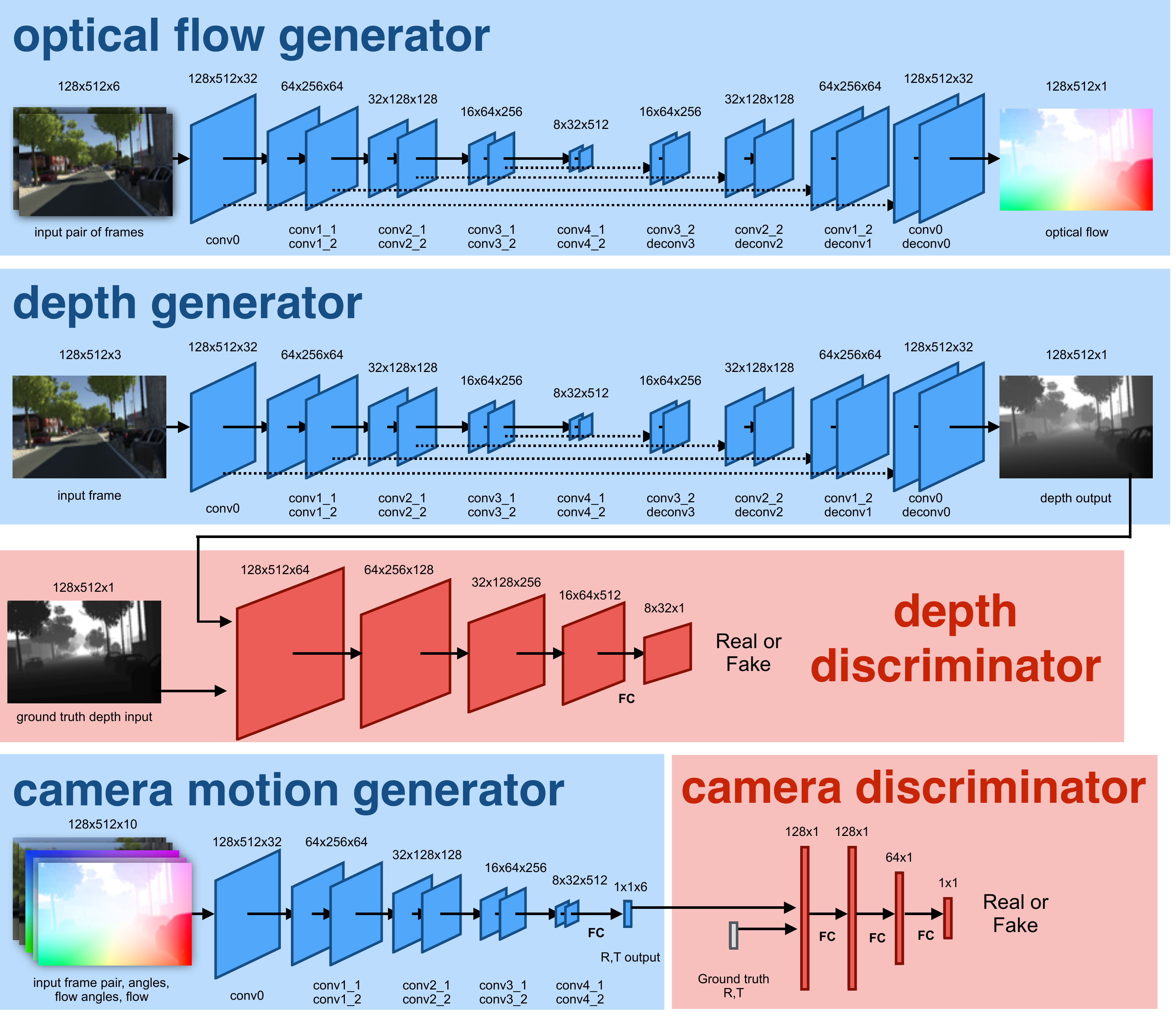} \\
    \end{tabular}
\centering
    \label{fig:architectures2}
    \caption{\textbf{Generator and discriminator architectures for  Structure from Motion.} Dashed lines indicate skip connections. }
\end{figure}

\subsection{Structure from Motion}
Our generator networks for the task of structure from motion is illustrated in Figure~\ref{fig:architectures2}. It includes three encoder-decoder convolutional networks with skip connections, which solve for optical flow, depth, and camera motion. The egomotion network uses RGB, optical flow and an angle field as input, and estimates the camera motion in SE(3). 
The depth network takes an RGB image as input and predicts logdepth. 
The depth discriminator consists of four convolution layers with batch normalization on the second and the third layers, and leaky ReLU activation after each layer. The depth  discriminator is fully convolutional as we are interested in the realism of every depth patch, as opposed to the depthmap as a whole. 

The egomotion discriminator is a 3-layer fully-connected network that takes $\{R,T\}$ matrices as input. The hidden layers have 128, 128, and 64 neurons, respectively, with batch normalization and a leaky ReLU after each layer.

\textbf{Stabilizing training.} 
In order to  make sure that generators and discriminators progress together during training, we update the generator  only when the discriminator has low enough loss. We add an updating heuristic such that if the likelihood loss of the discriminator is above a threshold $\theta_d,$ we do not update the generator. While discriminator is strong enough (below this threshold) and the generator is relatively weak (below a different threshold $\theta_g$), we update the generator twice 
in the iteration. 
In the experiments, we set $\theta_d$ to 0.695 and $\theta_g$ to 0.75.

\begin{figure*}[t]
    \centering
    \includegraphics[width=\linewidth]{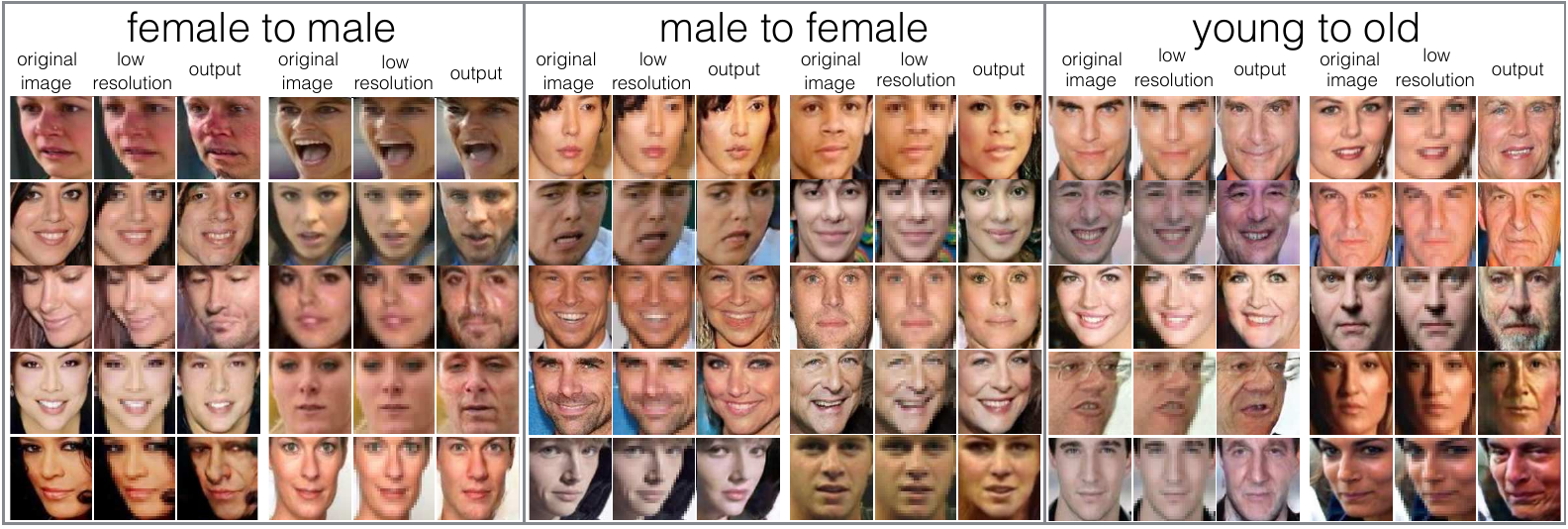}
     \centering
    \caption{AIGN on \textbf{gender transformation (female to male, male to female)} and \textbf{age transformation (young to old).}}
    \label{fig:more-male-merge}
\end{figure*}

\begin{figure}[t]
    \centering
    \includegraphics[width=0.8\linewidth]{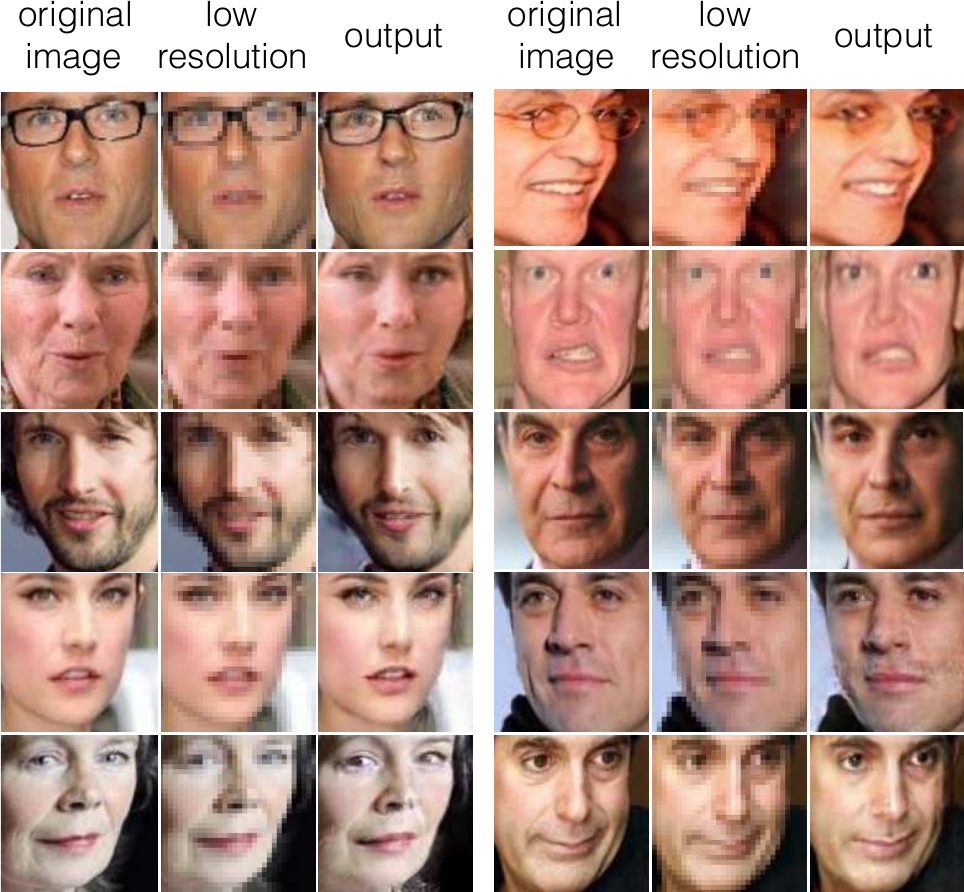}
     \centering
    \caption{AIGN on \textbf{age transformation (old to young).}}
    \label{fig:more-young}
\end{figure}

\subsection{Image Super-Resolution}

In Figure~\ref{fig:architectures}, we show the architecture of the generator and discriminator for image super-resolution.
The input image is first passed through a convolutional layer with 64 channels, then $n$ ``residual blocks''. Each residual block consists of two convolutional 
layers, with a batch normalization after each convolution layer and ReLU activation
after the first batch normalization. The output from the last block is further passed to two deconvolution layers
and generates the final image. 
The discriminator for this task consists of five convolution layers that use leaky ReLU activations and batch normalization, and one fully-connected layer that outputs one final value.
In all experiments, we use Adam optimizer with learning rate 0.0001.
All parameters are initialized with truncated normal distribution with variance 0.02.

\begin{figure}[t]
    \centering
    \includegraphics[width=0.95\linewidth]{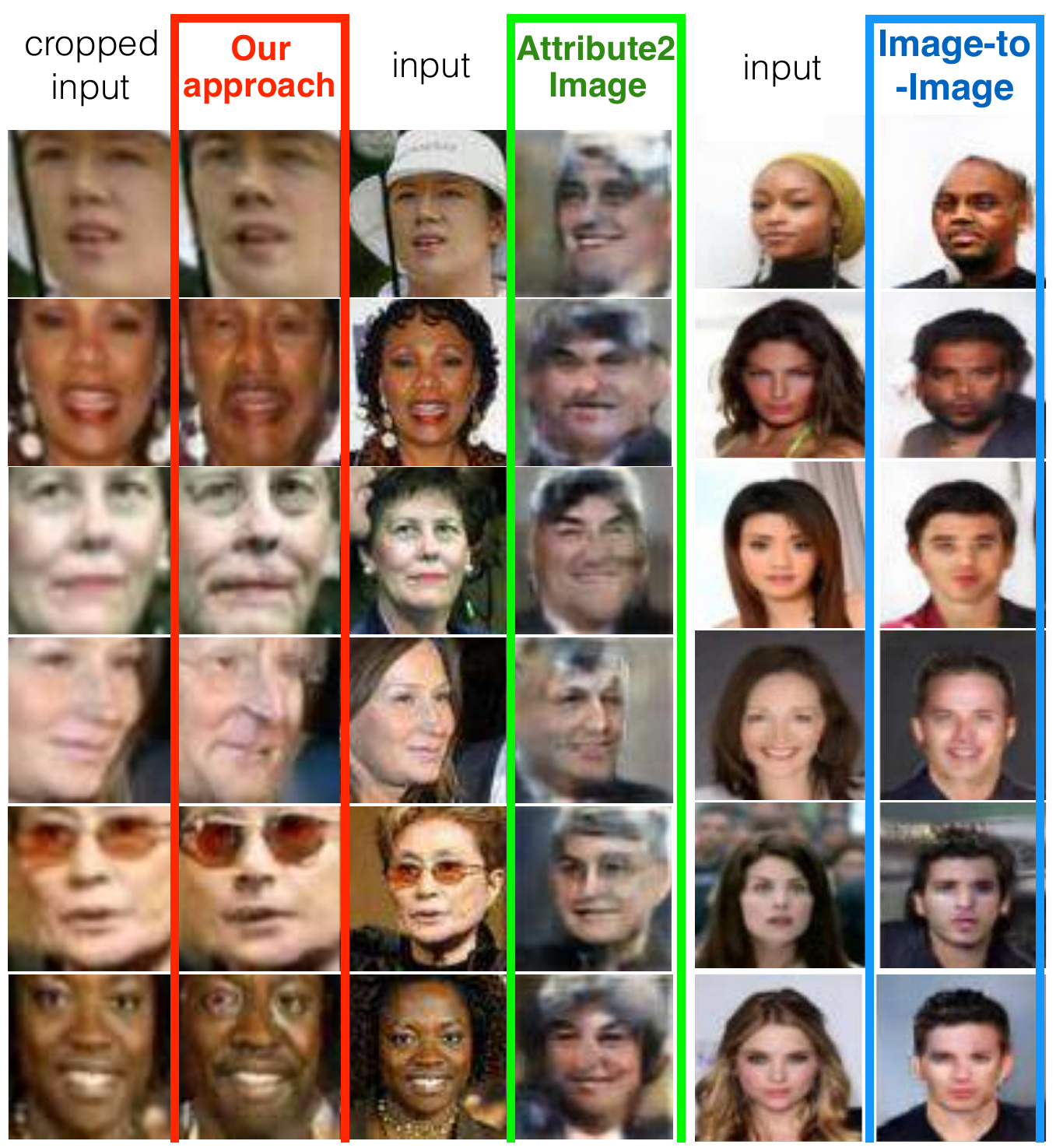}
     \centering
    \caption{Comparison with Attribute2Image \cite{DBLP:journals/corr/YanYSL15} and Unsupervised Image to Image Translation \cite{DBLP:journals/corr/DongNWG17} on \textbf{Gender transformation (female to male)}. Input to our model is a tight crop around the face, tighter than the crop used by \cite{DBLP:journals/corr/YanYSL15}. The proposed AIGN (\textit{Column 2}) provides more realistic results that better preserves the ``identity" of the subject while changing its gender, in comparison to previous work \cite{DBLP:journals/corr/YanYSL15} (\textit{Column 4}). We further show gender transformations from the model of \cite{DBLP:journals/corr/DongNWG17} (\textit{Columns 5,6}) where as we see the identity preservation is much weaker. Code is not available so we just paste some results from their paper. }
    \label{fig:baseline-male}
\end{figure}

\begin{figure}[t]
    \centering
    \includegraphics[width=0.9 \linewidth]{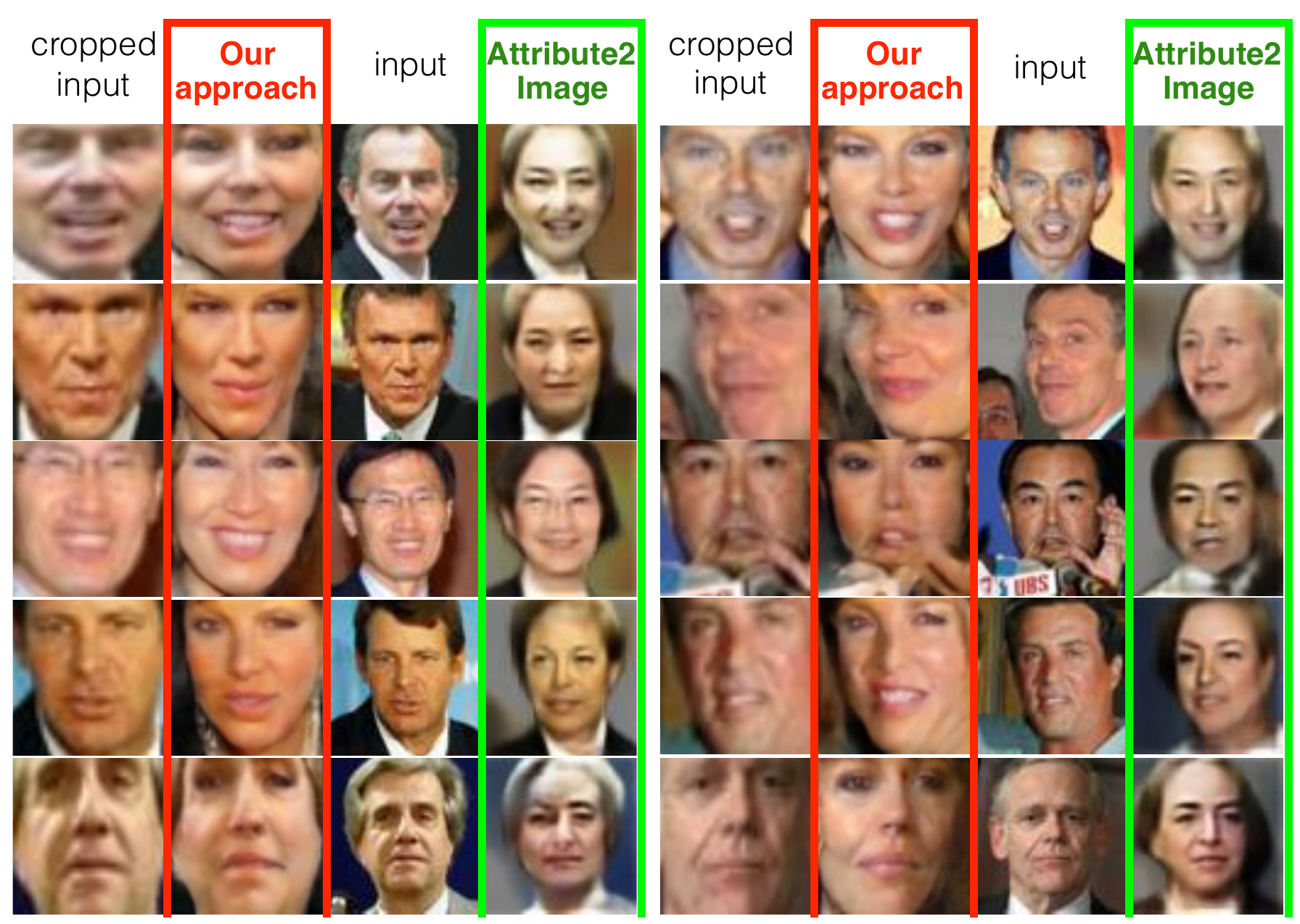}
     \centering
    \caption{Comparison of AIGN with Attribute2Image \cite{DBLP:journals/corr/YanYSL15}  on \textbf{gender transformation (male to female)}.}
    \label{fig:baseline-female}
\end{figure}

\begin{figure}[t]
    \centering
    \includegraphics[width=0.45\linewidth]{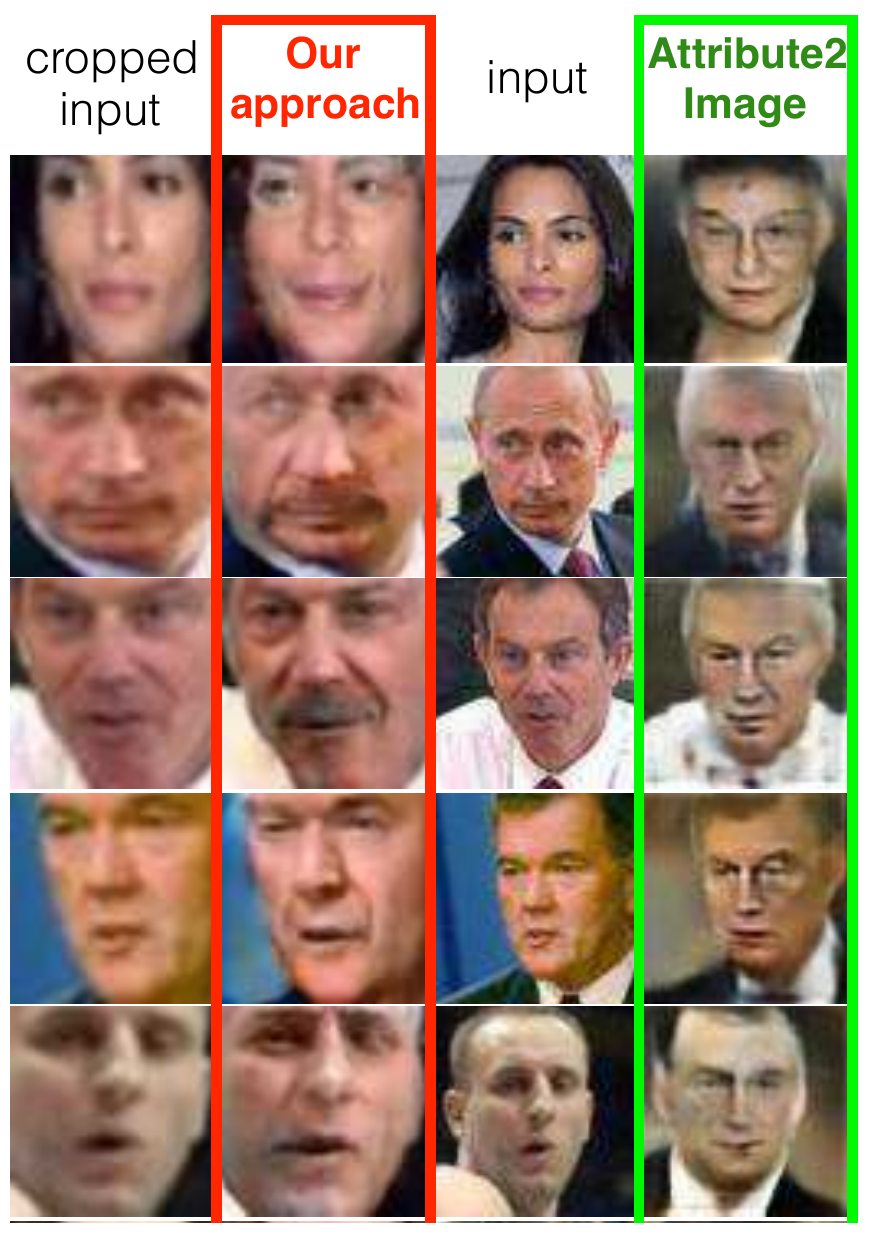}
    \includegraphics[width=0.45\linewidth]{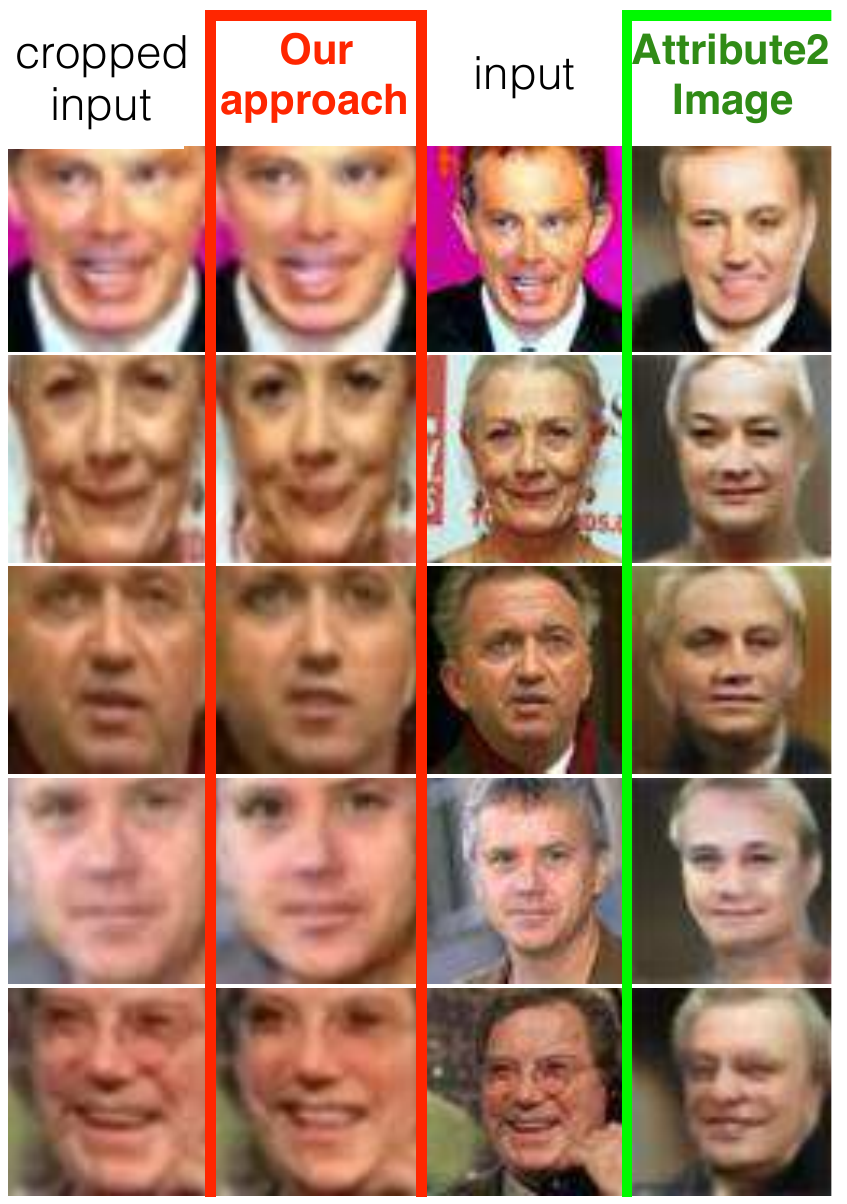}
     \centering
    \caption{Comparison of AIGN with Attribute2Image \cite{DBLP:journals/corr/YanYSL15} on \textbf{age transformation (left: young to old; right: old to young).}}
    \label{fig:baseline-old}
\end{figure}

In Figures~\ref{fig:baseline-male}, \ref{fig:baseline-female} and \ref{fig:baseline-old}, we compare our model with Attribute2Image  \cite{DBLP:journals/corr/YanYSL15} and with Unsupervised Image Translation \cite{DBLP:journals/corr/DongNWG17} for gender and age transformations. 
We use the code provided by the authors for our comparisons. 
In Figures~\ref{fig:more-male-merge} and \ref{fig:more-young},
we show additional results of our model on gender and age transformation.

\subsection{Inpainting}

Figure~\ref{fig:architectures} illustrates the architecture of our generator and discriminator for image inpainting.
The occluded input image and the corresponding mask are separately passed through two convolution 
layers, and then concatenated. The concatenated outputs are then passed to three deconvolutional layers  
to generate the inpainted image.  
The discriminator  consists of four convolutional layers  with leaky ReLU and batch normalization layers, and one fully connected layer that outputs one final value.  
In all experiments, we use the Adam optimizer, with a learning rate $1\mathrm{e}{-4}$
All parameters are initialized from the truncated Normal distribution, with variance 0.02.

\begin{figure}[t]
    \centering
    \includegraphics[width=1.0\linewidth]{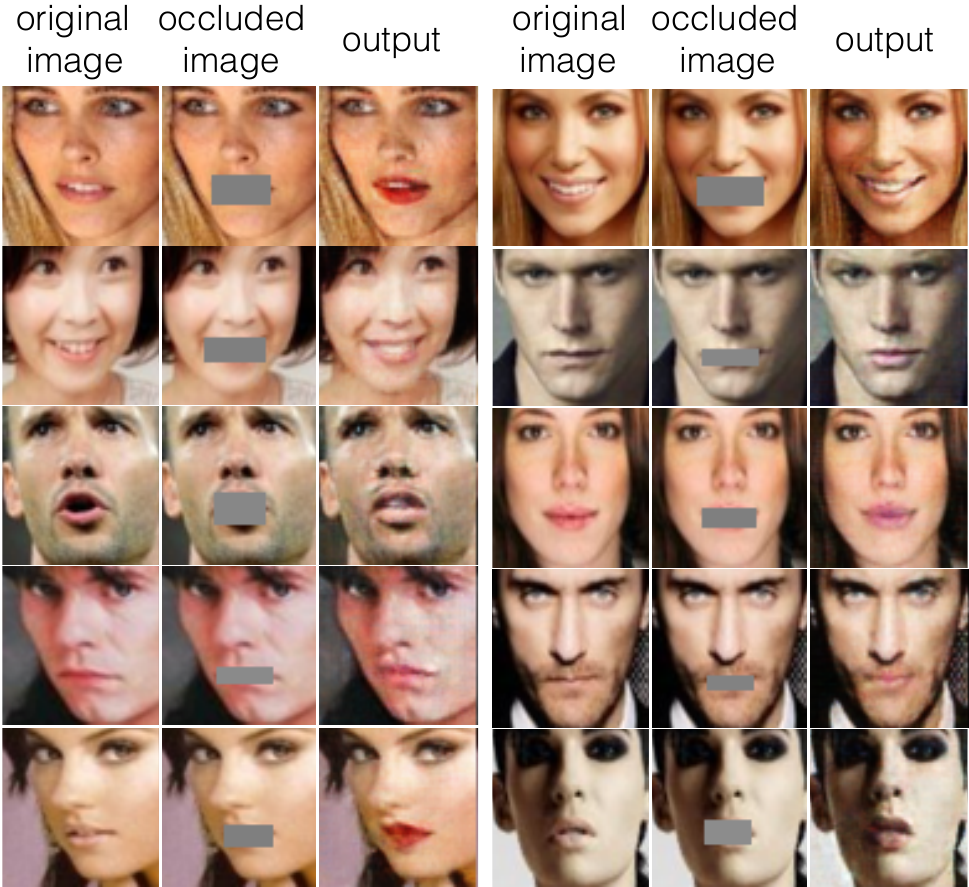}
     \centering
    \caption{Additional results  of AIGN on \textbf{biased image inpainting (big lips).}}
    \label{fig:more-lips}
\end{figure}

In Figure~\ref{fig:more-lips}, 
we show additional results on biased inpainting for making bigger lips.

%


\end{appendices}
\putbib[refs]
\end{bibunit}

\end{document}